\documentclass[10pt,twocolumn,letterpaper]{article}

\usepackage{times}
\usepackage{epsfig}
\usepackage{graphicx}
\usepackage{amssymb}

\usepackage[fleqn]{amsmath}
\usepackage{mathrsfs}
\usepackage{array}
\usepackage{graphicx}
\usepackage{color,soul}
\usepackage[inline]{enumitem}
\usepackage{tabularx}
\newcolumntype{M}[1]{>{\centering\arraybackslash}m{#1}}
\newcolumntype{C}{>{\centering\arraybackslash}m{.8cm}}
\usepackage{threeparttable}
\usepackage{cleveref}
\usepackage{caption,subcaption}
\usepackage{bm}

\usepackage{sectsty}

\sectionfont{\fontsize{12}{12}\selectfont}
\subsectionfont{\fontsize{11}{11}\selectfont}

\begin{document}

%%%%%%%%% TITLE
\title{3D Point Cloud Classification and Segmentation using 3D Modified Fisher Vector Representation for Convolutional Neural Networks}

\author{
Yizhak Ben-Shabat \\
Mechanical Eng. Dept.\\
Techion IIT \\
Haifa, Israel \\
{\tt\small sitzikbs@gmail.com}
\and
Michael Lindenbaum \\
Computer Science Dept.\\
Techion IIT \\
Haifa, Israel \\
{\tt\small mic@cs.technion.ac.il}
\and
Anath Fischer\\
Mechanical Eng. Dept.\\
Techion IIT \\
Haifa, Israel \\
{\tt\small meranath@technion.ac.il}
}
\date{}

\maketitle

%%%%%%%%%%%%%%%%%%%%%%%%%%%%%% Abstract %%%%%%%%%%%%%%%%%%%%%%%%%%%%%%%%%%%%%%
\begin{abstract}
   %\vspace*{12pt}
   \it
The point cloud is gaining prominence as a method for representing 3D shapes, but its irregular format poses a challenge for deep learning methods. The common solution of transforming the data into a 3D voxel grid introduces its own challenges, mainly large memory size. In this paper we propose a novel 3D point cloud representation called 3D Modified Fisher Vectors (3DmFV). Our representation is hybrid as it combines the discrete structure of a grid with continuous generalization of Fisher vectors, in a  compact and computationally efficient way.  Using the grid enables us to design a new CNN architecture for point cloud classification and part segmentation. In a series of experiments we demonstrate competitive performance or even better than state-of-the-art on challenging benchmark datasets.    
%\vspace*{10pt}
\end{abstract}

%\keywords{Deep Learning on 3D Point Clouds \and 3D Point Cloud Classification}

%%%%%%%%%%%%%%%%%%%%%%%%%%%%%% Sections %%%%%%%%%%%%%%%%%%%%%%%%%%%%%%%%%%%%%%
%The sections are saved in separate .tex files for easy access and ordering. additional sections may be added as needed
\section{Introduction}
\label{Sec:intro}

%Direct acquisition of 3D geometrical data using, e.g., LiDAR and RGBD cameras, providing point clouds, is effective, reliable and accurate. When applicable, it is usually superior to indirect methods such as stereo imaging. It is therefore becoming the preferred sensing approach for many applications such as modeling tools and autonomous systems.
 
Point clouds are commonly used for representing the sensory data associated with 3D objects and scenes. Recent sensing technologies make them more reliable and accurate and they are already in use in many applications such as modeling tools and autonomous systems.

We propose a new approach for analyzing a point cloud obtained by direct 3D sensors, using deep neural networks (DNNs), and especially convolutional neural networks  (ConvNets). ConvNets have shown remarkable performance in image analysis. Adapting them to point clouds is not, however, straightforward. ConvNets are built for input data arranged in fixed size  arrays, on which linear space invariant filters (convolutions) may be applied.  Point clouds, unfortunately, are unstructured, unordered, and contain a varying number of points. Therefore, they do not fit naturally into a spatial array (grid).  

Several methods for extending ConvNets to 3D point cloud analysis have been proposed \cite{maturana2015voxnet, wu20153d, qi2016volumetric}. A common approach is to  rasterize the 3D point data into a 3D voxel grid (array), on which  ConvNets can be easily applied. This approach, however, suffers from a tradeoff between its computational cost and its approximation accuracy.  We discuss this approach, as well as other common  point cloud representations, in Section \ref{Sec:related-work}. 

We take a different approach and use a new point cloud hybrid representation, the 3D Modified Fisher Vector (\textit{3DmFV}). 
The representation describes points by their deviation from a Gaussian Mixture Model (GMM). It has some similarity to the Fisher Vector representation (FV) \cite{perronnin2007fisher, sanchez2013image} but modifies and generalizes it in two important ways: the proposed GMM is specified using a set of uniform Gaussians with centers on a 3D grid, and the components characterizing the set of points, that, for Fisher vectors, are {\em averages}  over this set, are generalized to other functions of this set.

The representation is denoted hybrid because it combines the discrete structure of a grid with the continuous nature of the components.  It has several advantages. First, because it keeps the continuous properties of the point cloud, it retains some of the point set's fine detail and, under certain conditions, is lossless and invertible (in principle), and is therefore equivalent to featureless input. Second, the grid-like structure makes it possible to use ConvNets, which yields excellent classification accuracy even with low resolutions (e.g. 8$\times$8$\times$8). Finally, each component of the proposed representation represents a clear and meaningful property. 

The main contributions of this work are: 
\begin{itemize}
  \item We introduce a new hybrid representation for 3D point clouds (3DmFV) which is structured and order independent.
  \item We design a new deep ConvNet architecture (3DmFV-Net)  based on this % e 3DmFV 
  representation and use it for point cloud classification, obtaining state of the art results. 
  \item We conduct a thorough empirical analysis on the stability of our method.
  \item We extend the 3DmFV-Net to part segmentation of point clouds, and achieve state of the art results as well.  
\end{itemize}
We first review related work on 3D classification, part segmentation, and the FV representation in Section \ref{Sec:related-work}. Then, in Section \ref{Sec:method} we introduce and discuss the 3DmFV representation and 3DmFV-Net architecture. The classification and part segmentation results are presented in Section \ref{Sec:results}. Finally, we summarize in Section \ref{Sec:summary}.

\section{Related Work}
\label{Sec:related-work}

    \subsection{Deep learning on 3D data}
    \label{SubSec:DL_3D}
    
    \textbf{Point cloud features -} Handcrafted features for point clouds have shown adequate performance for many tasks. They can be divided into two main groups: local descriptors \cite{rusu2009fast, johnson1999using, tombari2010unique, guo2013rotational} and global descriptors \cite{aldoma2012our, wohlkinger2011ensemble, marton2011combined}. The comprehensive performance evaluation in  \cite{guo2016comprehensive} suggests guidelines for feature selection. However, optimal feature selection remains  non-trivial and highly data specific.
    
    \textbf{Deep learning on 3D representations - }
	3D data is commonly represented using one of the following representations:
	\begin{enumerate*}[label={(\alph*)}]
	  %star is for inline enumerate
	  	\item Multi-view,	
		\item Volumetric grid,
		\item Mesh,
		\item Point clouds. 
	\end{enumerate*}
Each representation requires a different approach for modifying the data to the form required by deep learning methods. 

Rendering 2D images of a 3D object from multiple views, as in \cite{qi2016volumetric}, transforms the learning domain from 3D to the well-researched 2D domain. Some information is lost in the projection process, but using multiple projections partially compensates. 

The volumetric, voxelized,  representation discretizes 3D space similarly to an image discretizing a camera projection plane. This  enables  a  straightforward extension of learning using 3D CNNs \cite{maturana2015voxnet,wu20153d, qi2016volumetric}. A volumetric representation is associated with a quantization tradeoff: choosing a coarse grid leads to quantization artifacts and to substantial loss of information, whereas choosing a fine grid 
% is associated with a very large 
significantly increases the number of voxels, which are mostly empty but still induce a high computational cost. Usually a grid size of 32$\times$32$\times$32 is chosen. 
A recent improvement applies an ensemble of very deep networks that extend the principles of Inception \cite{szegedy2017inception} and Resnet \cite{he2016deep} to voxelized data \cite{brock2016generative}, thus achieving the highest accuracy to date at the price of high computational cost and training time of weeks.

For the mesh representation, Spectral ConvNets \cite{bruna2013spectral, masci2015geodesic, boscaini2015learning} and anisotropic ConvNets \cite{boscaini2016learning} can be applied. These approaches utilize mesh topological structure, which is not always available. 

The point cloud representation is challenging because it is both unstructured and point-wise unordered. To overcome these challenges, the PointNet approach \cite{Qi_2017_CVPR,qi2017pointnet++} applies a symmetric function that is insensitive to the order, on a high-dimensional representation of the individual points. The Kd-Network \cite{klokov2017escape} imposes a kd-tree structure on the points and uses it to learn shared weights for nodes in the tree. Here we propose the 3DmFV representation, which is directly linked to the point cloud representation but can be used as input to a CNN.

\textbf{Part segmentation of 3D point clouds} - The objective here is to assign a label, for each point, corresponding to a semantically meaningful part of the shape, e.g., chair back or chair seat. The ShapeNet dataset, introduced in \cite{yi2016scalable}, was used in \cite{Qi_2017_CVPR} to extend the PointNet approach from point cloud classification to point-wise classification (part segmentation) by concatenating high-dimensional learned global and local features. The Kd-Network architecture was extended to part segmentation by mimicing an encoder-decoder architecture with skip-connections \cite{klokov2017escape}. 
% inspired by \cite{long2015fully}. 
Here, we extend the 3DmFV classification network to  perform part segmentation using locally and globally learned features.

    \subsection{Fisher vectors}
    \label{SubSec:FV}
	Before the age of deep learning, the bag of visual words (BoV) \cite{csurka2004visual} was a popular choice for image classification tasks. It extracted a set of local descriptors and assigned each of them to the closest entry in a codebook (visual vocabulary), leading to a histogram of occurrences. Perronin and Dance \cite{perronnin2007fisher}, and Perronnin and Thomas \cite{perronnin2010improving} proposed an alternative descriptor aggregation method, called the Fisher Vector (FV), based on the Fisher Kernel (FK) principle of \cite{jaakkola1999exploiting}. The FV characterizes data samples of varying sizes by their deviation from a generative model, in this case a Gaussian Mixture Model (GMM). It does so by computing the gradients of the sample's log-likelihood w.r.t. the model parameters (i.e., weight, mean and covariance). FV can be viewed as a generalization of the BoV as the histogram is closely related to the derivative w.r.t. the weight. Furthermore, it was shown in \cite{jaakkola1999exploiting} that, when the label is included as a latent variable of the generative model, the FK is asymptotically as good as the maximum a posteriori (MAP) decision rule for this model. The FV representation optimality  and independence of sample size make it a natural choice for representing point cloud data.
%  in the context of a deep learning approach for classification. 
	
In the context of image classification, the combination of FVs and DNNs was already considered \cite{simonyan2013deep, perronnin2015fisher}. A network composed of Fisher layers was suggested in \cite{simonyan2013deep}. Each layer performs semi-local FV encoding (on dense handcrafted features) followed by a dimensionality reduction, spatial stacking, and normalization. 
% The authors of \cite{sydorov2014deep} proposed to improve on FV using a joint learning framework of both the SVM classifier and the GMM visual vocabulary as a task-specific representation. 
A network composed of unsupervised and supervised layers was proposed in  \cite{perronnin2015fisher}. The unsupervised layers calculate features, FVs, and reduce their dimension. They are followed by supervised fully connected layers, trained with back propagation. 

The proposed 3DmFV shares some properties with the representation types above. Like the volumetric approach, it is based on a grid, but not a grid of voxels. It thus maintains the grid structure, which makes it a convenient input to a ConvNet, but it suffers less from quantization. Like the PointNet approach, its features are symmetric functions, making them order and structure independent. 
The architecture we propose, like that of  \cite{perronnin2015fisher}, combines unsupervised and supervised layers. However, it relies on the spatial properties of point clouds, which enables the use of ConvNets, substantially improving its performance.

%   The Fisher Vector representation of image descriptors was a popular choice for bag of visual words (BOV) methods and was superseded by deep learning methods. Perronin et al. \hl{*Micha - which perronin et al paper is best to cite here} **cite** proved optimality of this representation in a classification framework as it is equivalent to a generative Bayesian classification decision.  Therefore, it seems a natural choice for representing point cloud data.
%    In the FV representation, the high dimensional GMM parameters were iteratively computed using an expectation maximization (EM) method. In our pFV representation we position the Gaussians on a 3D lattice which nullifies some of FV's properties but also provides a more intuitive representation and empirically performs better for the classification task.  

\section{The 3DmFV-Net}
\label{Sec:method}
\begin{figure*}
		\centering
		\includegraphics[width=0.95\textwidth]{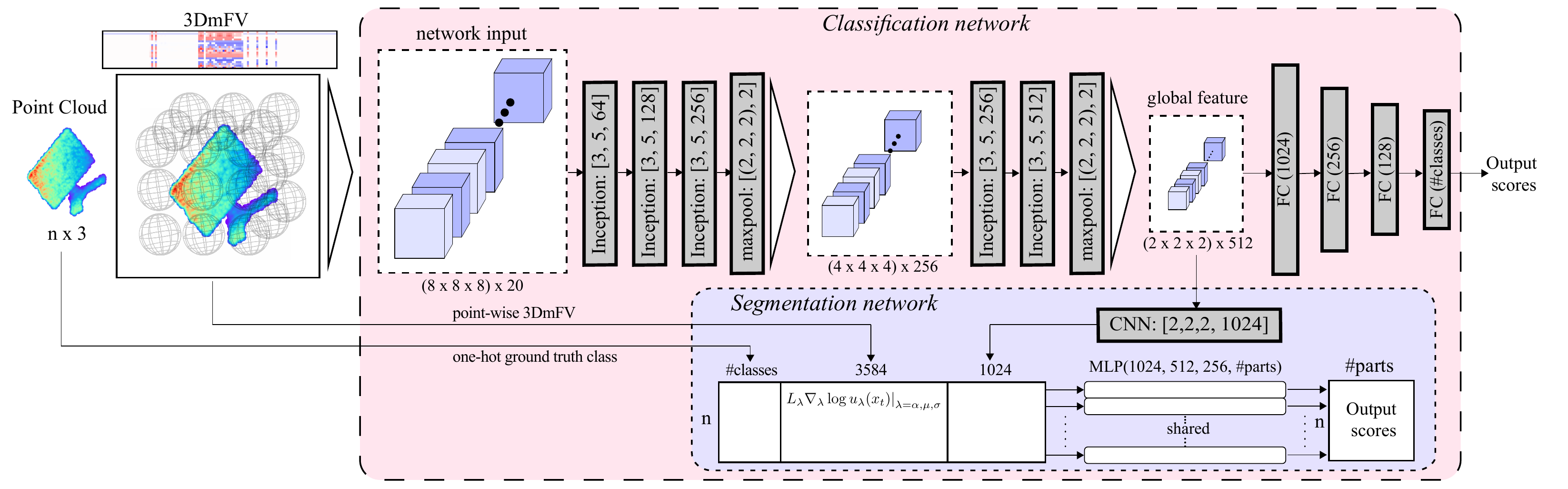}
		\caption{3DmFV-Net architecture}
		\label{fig:3DmFV_Net}	
	\end{figure*}

	The proposed 3DmFV-Net classification architecture consists of two main modules. The first converts an input point cloud to the 3D modified Fisher vector (3DmFV) representation and the second processes it in a CNN architecture. These main modules are illustrated in Figure \ref{fig:3DmFV_Net} and described below. 
	
\subsection{Describing point clouds with Fisher vectors}
\label{SubSec:FV2}
    
The proposed representation builds on the well-known Fisher vector representation. We start by formally describing the Fisher vectors (following the formulations and notation of \cite{sanchez2013image}) in the context of 3D points, and then discuss some of their  properties that lead to the proposed generalization and make them attractive as input to deep networks.

The Fisher vector is based on the likelihood of a set of vectors associated with a Gaussian Mixture model (GMM). Let $X=\{\bm{p}_t \in \mathbb{R}^3, t=1,...T\}$ be the set of 3D points of a given point cloud, where $T$ denotes the number of points in the set. Let $\lambda$ be the set of parameters of a $K$ component GMM $\lambda=\{(w_k,\mu_k,\Sigma_k), k=1,...K\}$, where $w_k,\mu_k,\Sigma_k$ are the mixture weight, expected value, and covariance matrix of \mbox{$k$-th} Gaussian. The likelihood of a single 3D point (or vector) $p$ associated with the $k$-th Gaussian density is 
    
    \begin{equation}
    	u_k(\bm{p}) = \frac{1}{(2\pi)^{D/2}|\Sigma_k|^{1/2}}\exp\left\{-\frac{1}{2}(\bm{p}-\mu_k)'\Sigma_k^{-1}(\bm{p}-\mu_k)\right\}.
    \end{equation}
    
%We follow \cite{perronnin2007fisher} and assume that the covariance matrices ($\Sigma_k$) are diagonal since a weighted sum of Gaussians with diagonal matrices can approximate any distribution with an arbitrary precision at lower computational cost than full covariance matrices. \MIC{rephrase}
The likelihood of a single point associated with the GMM density is therefore:
    \begin{equation}
    	u_\lambda(\bm{p}) = \sum_{k=1}^{K}w_ku_k(\bm{p}). 
    \end{equation}
    
Given a specific GMM, and under the common independence assumption, the Fisher vector, $\mathscr{G}_\lambda^X$, may be written as the sum of normalized gradient statistics, computed here for each point $\bm{p}_t$: 
%Every possible GMM (containing $K$ Gaussians) may be represented as a point, lying on a manifold in the parameter space. Starting from a specific GMM, the Fisher vector, $\mathscr{G}_\lambda^X$, is defined as the tangent direction (or gradient, in the parameter space) to the GMM that maximizes the joint likelihood of the set of points. This direction may be written, under the common independence assumption, as the sum of normalized gradient statistics, computed here for each point $x_t$: 
    \begin{equation}
    \label{eq:Fisher}
    	\mathscr{G}_\lambda^X = \sum_{t=1}^TL_\lambda\nabla_\lambda\log u_\lambda(\bm{p}_t),
    \end{equation}
    where $L_\lambda$ is the square root of the inverse Fisher Information Matrix.  
%        
%\noindent    
%\textbf{Gradients formulation} -  
The following change of variables, from $\{w_k\}$ to $\{\alpha_k\}$, ensures that $u_\lambda(x)$ is a valid distribution and simplifies the gradient calculation \cite{krapac2011modeling}:  
    \begin{equation}\label{eq:wk}
    	w_k = \frac{exp(\alpha_k)}{\sum_{j=1}^{K}exp(\alpha_j)}.
    \end{equation}
The soft assignment of point $\bm{p}_t$ to Gaussian $k$ is given by: 
    \begin{equation}\label{eq:gamma}
    	\gamma_t(k) = \frac{w_k u_k(\bm{p}_t)}{\sum_{j=1}^{K} w_j u_j(\bm{p}_t)}.
	\end{equation}     
\\
The normalized gradients are: 
	 \begin{equation} \label{eq:dev_w}
	 	\mathscr{G}_{\alpha_k}^X = \frac{1}{\sqrt{w_k}} \sum_{t=1}^T(\gamma_t(k)-w_k),
	 \end{equation}
	 \begin{equation} \label{eq:dev_mu}
	 	\mathscr{G}_{\mu_k}^X = \frac{1}{\sqrt{w_k}} \sum_{t=1}^T \gamma_t(k) \left( \frac{\bm{p}_t-\mu_k}{\sigma_k} \right),
	 \end{equation}
	 \begin{equation} \label{eq:dev_sig}
	 	\mathscr{G}_{\sigma_k}^X = \frac{1}{\sqrt{2w_k}} \sum_{t=1}^T \gamma_t(k) \left[ \frac{(\bm{p}_t-\mu_k)^2}{\sigma_k^2}-1 \right].
	 \end{equation}
These expressions include the normalization by $L_\lambda$ under the assumption that $\gamma_t(k)$ (the point assignment distribution) is approximately sharply peaked \cite{sanchez2013image}. 
For grid uniformity reasons, discussed in Section \ref{SubSec:3DmFV_generalization}, we adopt the common practice and work with diagonal covariance matrices. 
The Fisher vector is formed by concatenating all of these components: 
	\begin{equation}
		\mathscr{G}_{FV_\lambda}^X = \left(   
		\mathscr{G}_{\alpha_1}^X,...,\mathscr{G}_{\alpha_k}^X,
		{\mathscr{G}_{\mu_1}^X}',...,{\mathscr{G}_{\mu_k}^X}',
		{\mathscr{G}_{\sigma_1}^X}',...,{\mathscr{G}_{\sigma_k}^X}'
		 \right).
	\end{equation}
To avoid the dependence on the number of points, the resulting FV is normalized by the sample size T \cite{sanchez2013image}:
\begin{equation} \label{eq:FV_norm_T}
\mathscr{G}_{FV_\lambda}^X \leftarrow \frac{1}{T}\mathscr{G}_{FV_\lambda}^X.
\end{equation} 

See \cite{sanchez2013image}  for derivations, efficient implementation, and more details.

\subsection{Advantages of Fisher vectors as inputs to DNNs} %  deep neural networks}
\label{SubSec:FV-DNN}

A Fisher vector, representing a point set, may be used as an input to a DNN. It is a fixed size representation of a possibly variable number of points in the cloud.  Its components are normalized sums of functions of the individual points. Therefore, FV representation of a point set is invariant to order, structure, and sample size. 

Using a vector of non-learned features instead of the raw data goes against the common wisdom of deep neural network users. The reason is that features usually select some of the raw data properties and may lose some important characteristics. This is true especially when the feature extraction involves discretization, as is the case with voxel based representation.  We argue, however, that the Fisher vector representation, being continuous on the point set, suffers less from this disadvantage. We shall give three arguments (not proofs) in favor of this claim. 

\noindent
\textbf{a. Equation counting argument -} Consider a  $K$ component GMM. A set of $T$ points, characterized by $3T$ scalar coordinates, is represented using $7K$ components of the Fisher vector, every one of which is a continuous function of the $3T$ variables. Can we find another point set associated with the same Fisher components? We argue that for $T < 7K/3$, the set of equations specifying unknown points from known Fisher components is over-determined and that it is likely that the only solution, up to point permutation, is the original point set. 
% In this case the Fisher components description is a lossless description of the point set.
% 
While this equation counting argument is not a rigorous proof, such claims are common for sets of polynomial equations and for points in general position. If the solution is indeed unique, then the Fisher representation is lossless and using it is equivalent to using the raw point set itself. 
% This is very different from multi-view or voxel based representations, where quantization and loss are inherent.  

\begin{figure}
		\centering
		\includegraphics[width=0.22\textwidth]{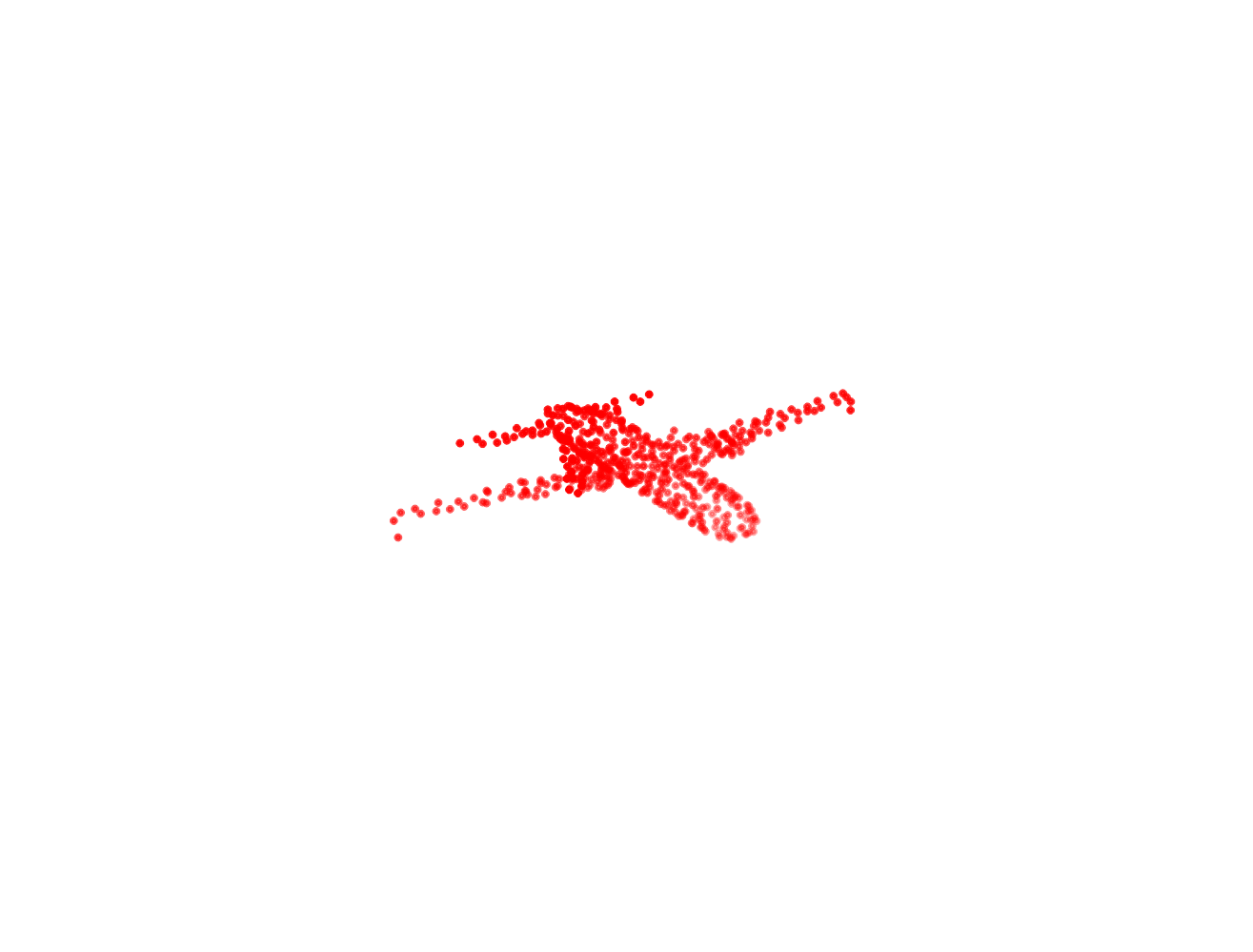}
		\includegraphics[width=0.22\textwidth]{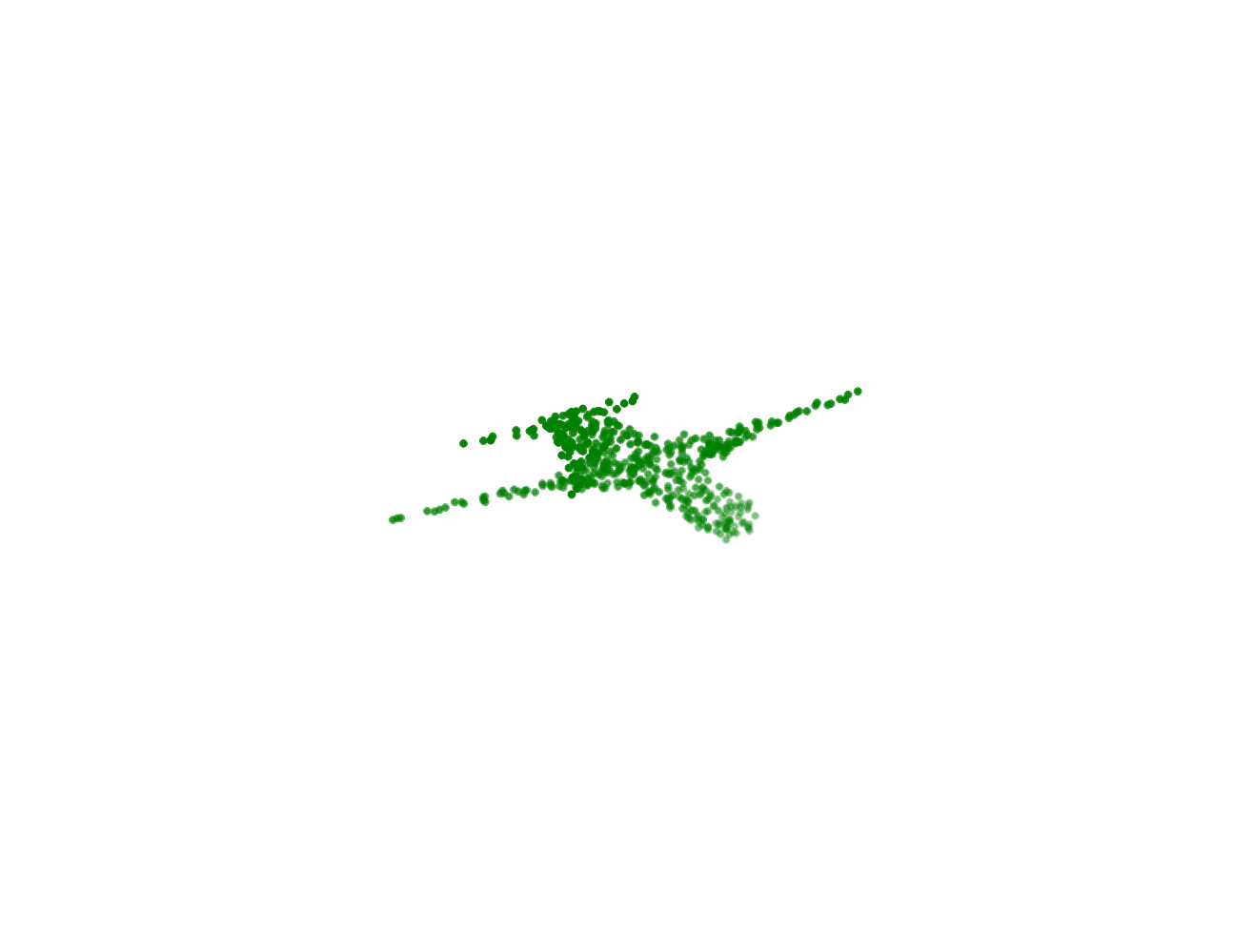}
		\caption{Point cloud reconstruction from FV representation using a deep decoder. The original (left), and the reconstructed point cloud (right).}
		\label{fig:3DmFV_decoder}	
	\end{figure}

\noindent
\textbf{b. Reconstructing the represented point structure  in simplified, isolated cases} 
% The represented point structure may be found from the Fisher components in simplified, isolated, cases: 
% To strengthen this intuitive claim, we now discuss three simple cases.
\\
\textbf{b.1 A single Gaussian representing a single point} -
Here,  $T=1$. By the sharply peaked  $ \gamma_t(k)$ assumption, there is only one Gaussian for which  $ \gamma_t(k) =1$. Inserting its FV components in Eq. \ref{eq:dev_mu} provides the point location: 

%	 \begin{equation} \label{eq:dev_mu_single_point_1}
%	 	\mathscr{G}_{\mu_k}^X =  \frac{x_1-\mu_k}{\sigma_k},
%	 \end{equation}

\begin{equation} \label{eq:x_single_point}
	 	\bm{p}_1 = \sigma_k\mathscr{G}_{\mu_k}^X +\mu_k.
\end{equation}
\\
%For further detail see Appendix \ref{appx:reconstruction_dev}.
%\textbf{A single Gaussian representing two point} -
%In this case $K=1$ and $T=2$. The Gaussian weight is $w_k = w_1 = 1$ and the soft assignment term becomes $\gamma_t(k) = \gamma_1(1) =\gamma_2(1) = 1$. Here we have six variables and six equations using \cref{eq:dev_w,eq:dev_mu,eq:dev_sig} where three of the equations are quadratic in $x$. Solving the equation system yields 
%\begin{equation} \label{eq:x_two_points}
%	 	x_{1,2} = \mu+\frac{\sigma\mathscr{G}_{\mu}^X}{2}\pm\sigma\sqrt{\frac{\mathscr{G}_{\sigma}^X}{\sqrt{2}}+1-\frac{\left(\mathscr{G}_{\mu}^X\right)^2}{4}}
%\end{equation}
% This represents eight possible solutions that reduce to four solutions due to point order invariance. For further detail see Appendix \ref{appx:reconstruction_dev}.
\textbf{b.2 A single Gaussian representing multiple points on one plane} -
We now show that, given a set of points sampled on a plane,  it is possible to reconstruct a plane from the FV representing the points.  The plane equation is  $ \hat{n}^Tp = \rho$, where $\hat{n} = (a, b, c)^T$ is the unit normal to the plane and $\rho$ is its distance from the origin. Using the assumption that $\gamma_t(k)$ is approximately sharply peaked \cite{sanchez2013image}, we consider the $k$-th Gaussian and the $T$ points for which  $\gamma_t(k) \approx 1$. For this Gaussian, eq. \ref{eq:dev_mu} is  simplified to: 
\begin{equation}
\mathscr{G}_{\mu}^X = \frac{1}{\sigma\sqrt{w}} \sum_{t=1}^T (\bm{p}_t - \mu).
\end{equation}
Changing the coordinate system to $x'y'z'$, for which the  origin is at the Gaussian center and an axis  $x'$ coincides with $\hat{n}$, leads to the following expression for the plane parameters (see Appendix \ref{appx:reconstruction_dev} %supplementary 
for a proof and illustrations) :
\begin{equation}
	a = \frac{\mathscr{G}_{\mu_x}}{\left\lVert \mathscr{G}_{\mu} \right\rVert},
	b = \frac{\mathscr{G}_{\mu_y}}{\left\lVert \mathscr{G}_{\mu} \right\rVert},
	c = \frac{\mathscr{G}_{\mu_z}}{\left\lVert \mathscr{G}_{\mu} \right\rVert},
	\rho = \sigma \left\lVert \mathscr{G}_{\mu} \right\rVert \sqrt{w} 
\end{equation}
%Here , $	\left\lVert \mathscr{G}_{\mu} \right\rVert = \sqrt{\mathscr{G}_{\mu_x}^2 + \mathscr{G}_{\mu_y}^2 + \mathscr{G}_{\mu_z}^2}$. See Appendix \ref{appx:reconstruction_dev_plane} for full detail. 
Objects are often approximately polyhedral, with each facet having at least one close Gaussian for which only this facet is close. This implies that such models may be reconstructed from the FV even for very large $T$. \\ 
% See appendix \ref{appx:reconstruction_dev} for simulation results. \\ 
%Reconstructing a plane in this way is meaningful for general reconstruction because, often, objects are approximately polyhedral, and for each Gaussian in a GMM there is only a single planar facet which influences it. See appensix \ref{appx:reconstruction_dev} for simulation results. \\ 
%In Section \ref{experiments} we demonstrate an approximate reconstruction of a point cloud from its (modified) FVs. \\
\textbf{c. Point cloud reconstruction from FV representation using a deep decoder} -
A direct expression for reconstructing a point cloud from its FV representation is not available for $K>1$. For illustration, we show now such a reconstruction obtained with a deep decoder, taking FV as an input and providing a point cloud. We consider a special FV, associated with a GMM with Gaussian centered on a grid; see Section \ref{SubSec:3DmFV_generalization} below. The decoder architecture is identical to the convolutional part of the classification network presented in Section \ref{SubSec:3DmFVNet_cls} followed by two fully connected layers:  FC($T$),FC($3T$).
 The loss function between the original and the reconstructed point sets, $S_1$ and  $S_2$, should be invariant to point order.  
We  use a loss function  that is the sum of Chamfer distance and the Earth mover's distance, which were used (separately) in \cite{fan2016point}.
%We use a loss function (\ref{eq:joint_loss}) which is the sum of the two cost functions proposed in \cite{fan2016point}: the chamfer distance (CD) (\ref{eq:chamfer_loss}), and the earth movers distance (EMD) (\ref{eq:emd_loss}). 
%\begin{equation} \label{eq:chamfer_loss}
%\begin{split}
%d_{CD}(S_1,S_2) = 
% \sum_{\bm{p}_{s_1} \in S_1}\min_{\bm{p}_{s_2} \in S_2} \lVert \bm{p}_{s_1} - \bm{p}_{s_2} \rVert_2^2 + \\
% \sum_{\bm{p}_{s_2} \in S_2}\min_{\bm{p}_{s_1}\in S_1} \lVert  \bm{p}_{s_1} - \bm{p}_{s_2} \rVert_2^2 
% \end{split}
%\end{equation}
%\begin{equation} \label{eq:emd_loss}
%d_{EMD}(S_1,S_2) = \min_{\Phi:S_1 \rightarrow S_2} \sum_{\bm{p}_{s_1} \in S_1} \lVert \bm{p}_{s_1}-\Phi(\bm{p}) \rVert_2 
%\end{equation}
%
%\begin{equation} \label{eq:joint_loss}
%L(S_1, S_2) = d_{EMD}(S_1,S_2) + d_{CD}(S_1,S_2)  
%\end{equation}
%Where $\Phi : S_1\rightarrow S_2$ is a bijection between two equal sized sets. 
Figure \ref{fig:3DmFV_decoder} shows a qualitative comparison between the original point cloud and the reconstructed point cloud. It shows that the decoder captured the overall shape while not positioning the points exactly in their original position.  

%The Fisher components and readily computable. Can we find another set $\bf X'$ associated with the same Fisher components ? Note that, by symmetry, any set obtained from $\bf X_0$ by point permutation is associated with the same Fisher components. We argue intuitively that when the original point set $\bf X_0$ consists of points in general position, then it is unlikely that we get other such $\bf X$ sets. Suppose that there is such a set $\bf X'$. Then, the functions $f_j(\bf X)$ of its $T$ points should match the Fisher components of $3T/7<K$ of the Gaussians in the mixture. This seems reasonable because the number of variables ($3T$) and the number of constraints are the same. However the fisher components calculated from the set $\bf X'$ should match those calculated from $\bf X_0$, and this seems possible only accidentally. 

%Indeed, Fisher vectors have already been used to represent images as an input to a deep network. It was shown that a deep network improves classification performance compared to linear classifier, but does not work better than a CNN working directly on the pixel values \cite{perronnin2015fisher}. \MIC{to be continue. add ref and describe it}\IBS{ref added. We addressed this in the new related work section so I think no further description is necessary. I dont think we need this paragraph anymore at all}

\subsection{Generalizing Fisher vectors to 3D modified Fisher vectors}
\label{SubSec:3DmFV}
    
We propose to generalize the Fisher vector along two directions: 

\textbf{Choice of the mixture model - }
Originally, the mixture model was defined as a maximum likelihood model. This makes the model optimally adapted to the training data and gives the Fisher representation of each Gaussian the nice property of being sensitive only to the deviation from the average training data. It is not the only valid option, however, and not a good choice if we prefer to maintain a grid structure. Therefore, in the proposed generalization we shall use other mixture models that rely on Gaussian grids.  

\textbf{Choice of the symmetric function - }  As apparent from eq. (\ref{eq:Fisher}), the components of the Fisher vector are sums over all input points, regardless of their order and any structure they create. They are {\em symmetric} in the sense proposed in \cite{Qi_2017_CVPR} and are therefore adequate for representing the orderless and structureless set of points. Note also that any other symmetric function, applied over the summands in eq. (\ref{eq:Fisher}), would also induce a vector that can represent orderless sets. We will propose such functions and use them instead of or in addition to the Fisher vector sums.
 
\subsection{The proposed 3DmFV generalization}
\label{SubSec:3DmFV_generalization}

\textbf{Changing the mixture model - }
For the underlying density model, we  use a mixture of Gaussians with Gaussian centers ($\mu_k$) on a uniform 3D 
$m\times m\times m$ grid. % structure
Such Gaussians induce a Fisher vector that preserves the point set structure: the presence of points in a specific 3D location would significantly influence only some, pre-known, Fisher components.
The other GMM parameters, weight and covariance, are common to all Gaussians. The weights are selected as $w_k = \frac{1}{K}$ and the covariance matrix as $\Sigma_k = \sigma_k I$ with $\sigma_k = \frac{1}{m}$. Recall that all points are contained in the unit sphere. 
The uniformity is essential for shared weight (convolutional) filtering. The size of the mixture model is moderate and ranges from $m=3$ to  $9$.

The proposed uniform model is not as effective as the maximum likelihood model for representing the distribution of point clouds. Recall, however, that the GMM does not represent a specific model or a specific class but rather the average model, which is much closer to uniform. The inaccuracy is more than compensated for by the power of the convolutional network, as we shall see in the comparison between the different models.  
 
\textbf{Changing/Adding other symmetric functions -} 
% Theoretically, to get the promised optimality, the Fisher vector components should be the expected values of the gradient components. Practically, ergodicity \IBS{? couldnt find translation} is used, and the expected value is replaced with an  average (or sum) over the input  points. 
For Fisher vectors, the sum is used as a symmetric function. 
While the sum is asymptotically optimal, it does not give full information about the input points for  finite point sets. For small point sets the Fisher vectors may be invertible, as suggested above, implying that the FVs implicitly carry the full information about the set. For the practical case of large finite point sets, we propose to add information. To maintain the order independence, other summarizing features should be symmetric as well. 

We experimented with several options for additional symmetric functions, and eventually chose the maximum and minimum functions. Note that the maximum was also used in \cite{Qi_2017_CVPR} as a single summarizing feature for each of the learned features. Thus, the components of the proposed generalized vector, denoted 3DmFV, are given in eq. \ref{eq:3DmFV_definition} and obtained as follows: each component is either a sum, max, or min function, evaluated on the set of one gradient component. In our experiments we found that partial, more compact, representations (especially those focusing on the minimum and maximum function) may lead to improved accuracy, and we describe them as well. However, we also found that the minimal weight derivative is always a constant and omitted this specific function. The minimal value associated with a specific Gaussian corresponds to the farthest point and its $\gamma_t(k)$ value, which is 0 in practice.

%The pFV representation heuristically concatenates minimum, maximum and sum of $\mathscr{G}_{\mu_i}^X$ and $\mathscr{G}_{\sigma_i}^X$ and maximum and sum gradients for $\mathscr{G}_{\alpha_i}^X$(since the minimum is often zero for all Gaussians):
        \begin{equation} \label{eq:3DmFV_definition}
    	\mathscr{G}_{3DmFV_\lambda}^X = \left[ \begin{array}{c} 
    	\left. \sum_{t=1}^TL_\lambda\nabla_\lambda\log u_\lambda(\bm{p}_t) \right|_{\lambda=\alpha,\mu,\sigma} \\
    	 \left. \max_t(L_\lambda\nabla_\lambda\log u_\lambda(\bm{p}_t)\right|_{\lambda=\alpha,\mu,\sigma} \\
    	 \left. \min_t(L_\lambda\nabla_\lambda\log u_\lambda(\bm{p}_t))\right|_{\lambda=\mu,\sigma} \end{array}\right]
    	\end{equation}
	
\begin{figure}
		\centering
		\includegraphics[width=0.45\textwidth]{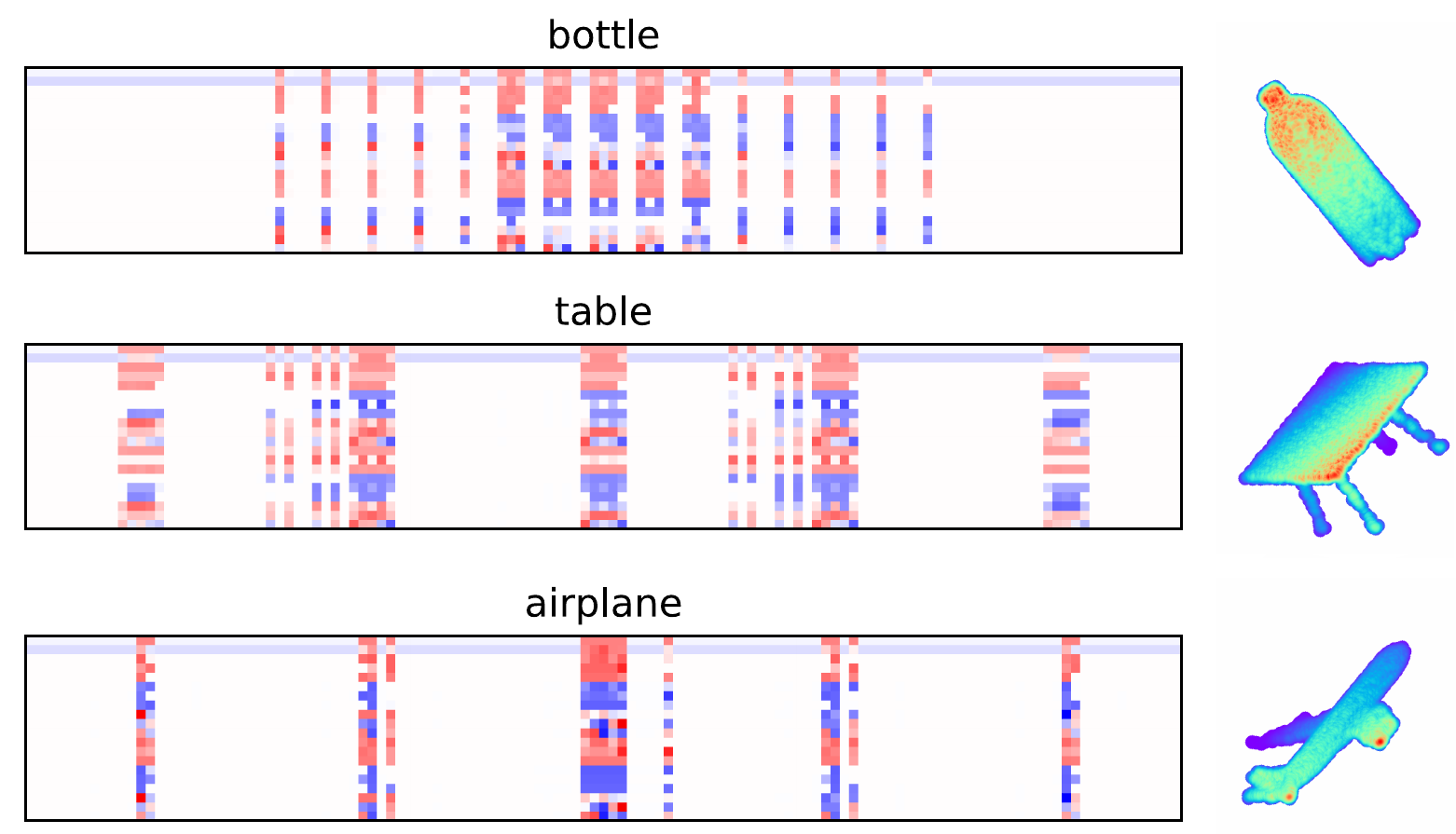}
		\caption{3DmFV representation (left) and the corresponding point cloud (right)}
		\label{fig:3DmFV_PC}	
	\end{figure}

For $K=m^3$ Gaussians, there are therefore $ 20\times K$ components in the 3DmFV representation ($20=3(3+3)+2$). The 3DmFV is best visualized as a $20\times K$ matrix. 
%  which can be easily reshaped into the spatial grid structure $m\times m\times m\times 20$ where $K = m^3, m \in \mathbb{Z}$.
%    
Figure~\ref{fig:3DmFV_PC} depicts a point cloud (right) and its 3DmFV representation ($m = 5$) as a color coded image (left). Each column of the image represents a single Gaussian in a 5$\times$5$\times$5 Gaussian grid. Zero values are white whereas positive and negative values correspond respectively to the red and blue gradients. Note that the representation lends itself to intuitive interpretation. For example, many columns are white, except for the first two top entries. These correspond to Gaussians that do not have model points near them; see eq. \ref{eq:dev_w}.

\noindent
\textbf{Normalization}
Following \cite{perronnin2010improving} (Sec. 2.3), we applied two  consecutive normalizations on the 3DmFV representation: First, we applied an element-wise signed square root normalization, and then an L2 normalization over all 20 vectors corresponding to all Gaussians and a single feature $\lambda_i$. This last normalization equalizes the derivatives with respect to different parameters. 

%
%as proposed in \cite{perronnin2010improving}: 
%    \begin{equation}
%    	\left[ \mathscr{G}_{PFV_\lambda}^X \right]_i \leftarrow sign \left( \left[\mathscr{G}_{PFV_\lambda}^X \right]_i\right) \sqrt{\left| \left[ \mathscr{G}_{PFV_\lambda}^X \right]_i \right|}
%    \end{equation}
%    Then, we apply an L2 normalization in order to limit the magnitude of the network input elements.
%    \begin{equation}
%    	\mathscr{G}_{PFV_\lambda}^X = \frac{\mathscr{G}_{PFV_\lambda}^X}{\sqrt{\mathscr{G}_{PFV_\lambda}^{X'} \mathscr{G}_{PFV_\lambda}^X}}
%    \end{equation}
%    
 
    \subsection{3DmFV-Net classification architecture}
    \label{SubSec:3DmFVNet_cls} 
\begin{figure}
			\centering
			\includegraphics[width=0.44\textwidth]{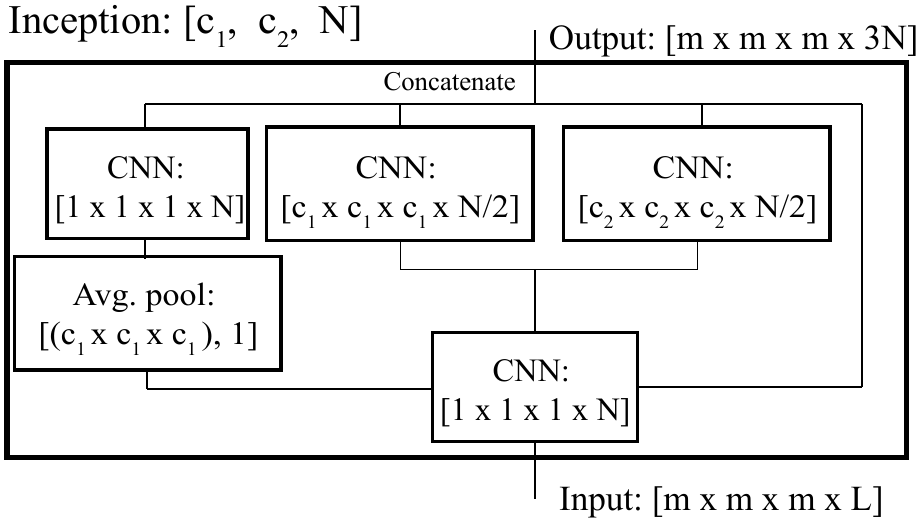}
			\caption{Inception module used in 3DmFV-Net}
			\label{fig:inception_module}	
		\end{figure}

    The proposed network receives a point cloud and converts it to a 3DmFV representation on a grid. The main parts of the network consist of an Inception module \cite{szegedy2015going}, visualized in Figure \ref{fig:inception_module}, maxpooling layers, and finally four fully connected layers. The network output consists of classification scores; see Figure \ref{fig:3DmFV_Net}. The network is trained using back propagation and standard softmax cross-entropy loss with batch normalization after every layer and dropout after each fully connected layer. The network has approximately 4.6M trained parameters, the majority of which are between the last maxpooling layer and the first fully connected layer.   
		
\subsection{3DmFV-Net part segmentation architecture}
    \label{SubSec:3DmFVNet_seg}
The 3DmFV-Net architecture is extended to perform part segmentation. We solve this problem using a per-point classification approach. The proposed architecture combines local geometrical information of each point with the global context of the entire point cloud. 
 The local information is encoded in the per-point 3DmFV representation (i.e., the derivative value of each GMM Gaussian with respect to $\lambda$ at that specific point), and the global information is added by concatenating the output of the convolutional part of the classification architecture (before the FC classifiers) to each point's 3DmFV representation. Additionally,  to fairly compare to other methods that assume a given label, we concatenate a one-hot representation of the ground truth label; however, this is not strictly necessary since ommitting the label only slightly reduces performance ($0.2\%$). These features are then aggregated into a combined representation using a multilayer perceptron with shared weights across points, similarly to \cite{Qi_2017_CVPR}.
  The details of the 3DmFV-Net segmentation architecture are illustrated in Figure \ref{fig:3DmFV_Net}.

\section{Experiments}
\label{Sec:results}
%In this section we evaluate the classification and part segmentation obtained by the proposed 3DmFV-Net, and compare them to other algorithms and to several variations of the proposed algorithm.
We first evaluate the classification performance of our proposed 3DmFV-Net and compare it to previous approaches. We then evaluate several variants of the proposed representation. Next, we analyze our method's robustness to noise. Finally, we evaluate the part segmentation results obtained by our method and compare them to the state of the art.
\subsection{Implementation}
 
\textbf{Datasets} - We evaluate all classification algorithms on the ModelNet40 dataset \cite{wu20153d}. It consists of  12311 CAD models from 40 object categories, represented as triangle mesh. The data is split into  a training set (9843 models) and a test set (2468). To generate point clouds, the mesh is sampled as described in \cite{Qi_2017_CVPR}. We also experimented with the ModelNet10 dataset, which contains 4899 CAD models from 10 object classes split into 3991  for training and 908 for testing. 

\textbf{Training details}: Unless otherwise specified, the 3DmFV-Net (Figure \ref{fig:3DmFV_Net}) was trained using an Adam optimizer with a learning rate of $0.001$ with a decay of $0.7$ every $20$ epochs. The point cloud (of 2K points)  is centered around the origin and scaled to fit a cube of edge length 2. Data is augmented using random anisotropic scaling (range: $[0.66, 1.5]$)  and random translation (range: $[-0.2, 0.2]$) in each axis, similarly to \cite{klokov2017escape}. We used Tensorflow on a NVIDIA Titan Xp GPU.  Training took $\sim 7$ hours.

\subsection{Classification performance}
%
%\textbf{Implementation -}
%We trained the algorithm in the standard way described above, and added some special measures which improved its performance. First, we used a specialized augmentation, \MIC{we should describe it. This is one innovation. no ?} \dots 
% \textbf{Benchmark classification performance}:
%Table.\ref{table:ClassificationAcc} compares the 3DmFV-Network with previous approaches. Clearly, the proposed method wins over most methods, and in particular, over the PointNet approach that is most similar to it \cite{xxx}. It is less accurate, though, than the more complex VRN-ensemble \MIC{This name is not in the table. Also please write a short description of this method}, which represents the cloud by a $32 \times 32 \times 32$ voxels. It uses an ensemble method, which averages 6 models, each trained for 6 days. We achieve comparable results to Kd-network \cite{xxx} which requires a very large input cloud ($\sim32K$ points\MIC{What happens with 2K points ? } in comparison to $\sim2K$ that we use\MIC{Where do we say that, before}), and is much less robust; see below.

Table \ref{table:ClassificationAcc_ModelNet} compares the 3DmFV-Net with previous approaches on the ModelNet40 and Modelnet10 datasets. Clearly, the proposed method wins over most methods. 
%, and in particular, over the PointNet approach that is most similar to it \cite{xxx}. 
It is comparable to the Kd-network, which requires a much larger input($\sim32K$ points) and is not very robust to rotations and noise \cite{klokov2017escape}. It is less accurate only than the more complex VRN ensemble method \cite{brock2016generative}, which operates on voxelized input and averages 6 models, each trained for 6 days. However, it is slightly better than a single VRN model. Note also that direct comparison is somewhat unfair because, unlike the voxelized description, point based methods (like ours) do not have direct access to the mesh. 
 
We also tested a combination of the 3DmFV representation ($m=8$) with the simpler convolutional network that mimics the one used in VoxNet \cite{maturana2015voxnet}. Although this combination (denoted 3DmFV+VoxNet)  uses a lower resolution grid than VoxNet ($32^3$ voxels), it is more accurate. Thus, the  performance boost of the 3DmFV-Net may be attributed to both the representation and the architecture.

    \begin{table}[tb]
	\centering	
		\tabcolsep = 0.01\textwidth %0.18cm
		\begin{tabular}{| m{0.17\textwidth} | M{0.11\textwidth} | M{0.11\textwidth}|}
		\hline
		 \centering\textbf{Method} & \centering\textbf{ModelNet10} & \centering\textbf{Modelnet40}
    	 \tabularnewline
   	 \hline
    	MVCNN \cite{su2015multi} & - & 90.1\\

        3DShapeNets \cite{wu20153d} & 83.5 & 77.32\\
        VoxNet \cite{maturana2015voxnet} & 92.0 & 83.0 \\
        VRN (One-View) \cite{brock2016generative} & - & 88.98\\
        VRN \cite{brock2016generative} & 93.6 & 91.33\\
        VRN ensemble \cite{brock2016generative} & 97.14 & 95.54\\
    	FusionNet \cite{hegde2016fusionnet} & 93.1 & 90.8\\    
    \hline
    	PointNet \cite{Qi_2017_CVPR} & - & 89.2\textsuperscript{a} \\
        PointNet++ \cite{qi2017pointnet++} & - & 90.7\textsuperscript{a} \\
        Kd-network \cite{klokov2017escape}& 94.0\textsuperscript{b} / 93.3\textsuperscript{a} & 91.8\textsuperscript{b} / 90.6\textsuperscript{a} \\
    \hline
    3DmFV+VoxNet & 94.3 & 88.5\textsuperscript{c} \\
    Our 3DmFV-Net & 95.2\textsuperscript{a}\textsuperscript{c} &91.6\textsuperscript{c}/91.4\textsuperscript{a}\\
    \hline
	\end{tabular}
	\caption{Classification accuracy on ModelNet40 and ModelNet40 datasets. The point based methods use \textsuperscript{a}1024,\textsuperscript{b}32768, \textsuperscript{c}2048 points.}

	\label{table:ClassificationAcc_ModelNet}
\end{table}

\subsection{Testing variations of the 3DmFV representation}

% The proposed 3DmFV representation is indeed powerful. 
%  The following set of experiments
We now test partial, more compact variants of the proposed representations. 
The first variant is the Fisher vectors. 
The 3DmFV generalization of FV uses different symmetric functions (and not only the sum, as in FV), and different GMMs. We considered the combinations of several symmetric functions, with GMMs obtained either from the maximum likelihood (ML) optimization obtained by the expectation maximization (EM) algorithm or from Gaussians on a 3D 5$\times$5$\times$5 grid. The tested symmetric functions include maximum (3DmFV-max), minimum (3DmFV-min), and sum of squares (3DmFV-ss). 
We tested these combinations with a nonlinear classifier (4 fully connected layers of sizes $(1024, 256, 128, 40)$ with ReLU activation). For reference to the original FVs, we tested the ML GMM with a linear classifier as well. All models were represented by 1024 points. Table \ref{table:FV_3DmFV_EM_grid} reveals that the 3DmFV representation always wins over the FV representation, and that using grid GMM is comparable to the optimal ML GMM. Using  maximum or minimum as a symmetric function yields comparable results with fewer parameters. Note that with the convolutional network, possible only with grid GMM, the accuracy is much higher ( Table \ref{table:ClassificationAcc_ModelNet}).

%\textbf{FV vs. 3DmFV} We evaluated the classification performance of 3DmFV compared to the original FV representation using a linear and a non-linear classifier. The linear classifier is a single fully connected layer with a neuron for each class (without activation) and the non-linear classifier is constructed of three fully connected layers of sizes $(1024, 256, 128)$, a ReLU activation and dropout after each layer followed by a fully connected layer with a neuron for each class.
%
%\noindent
%\textbf{EM vs. Grid} We evaluated two main Gaussian types. The first evaluates Gaussian parameters using an Expectation Maximization algorithm (EM) on 50\% of the training data.  The second places the Gaussians on an axis-aligned grid with the equal weights and a predefined standard deviation $\sigma = \frac{2}{m}$, here $m$ represents the number of Gaussians along an axis. Evaluation performed for 125 Gaussians and 1024 points for each model.
%
%In Table. \ref{table:FV_3DmFV_EM_grid} we show that the 3DmFV representation performs better than FV representation in all cases. In addition, surprisingly the grid Gaussians show similar results as EM learned Gaussians for 3DmFV and better results for FV. Therefore we can use the grid Gaussians to learn a more descriptive representation using CNN's.   

    \begin{table}[tb]
	\centering	
	\tabcolsep = 0.01\textwidth %0.18cm
	\begin{tabular}{| M{0.12\textwidth} | M{0.08\textwidth} | M{0.1\textwidth}| M{0.1\textwidth}|}   
		\hline
	 \centering\textbf{Rep.} & \centering\textbf{ML + LinCls }& \centering\textbf{ML + NonLinCls} & \centering\textbf{Grid + NonLinCls}
     \tabularnewline
     \hline
     FV & 58.4 & 82.8 & 84.5 \\
     3DmFV-ss & 58.8 & 85.0 & 84.4 \\
     3DmFV-min & 67.7 & 87.7 & 86.1\\
     3DmFV-max & 68.6 & 87.4 & 85.3\\
     3DmFV & 76.8 & 88.0 & 87.7 \\
     \hline
	\end{tabular}	
	\caption{Classification accuracy of the proposed 3DmFV and FV  representation computed on EM-learned Gaussians and Gaussians positioned on a grid using a linear and non-linear classifier. Note that convolution layers are not used in this experiment.}
	\label{table:FV_3DmFV_EM_grid}
\end{table}
\begin{table*}[tb]
\setlength\tabcolsep{1.5pt}
    \begin{tabular}{|l|C|CCCCCCCCCCCCCCCC|}
    \hline
     method&mean&aero&bag&cap&car&chair&ear phone&guitar&knife&lamp&laptop&motor bike&mug&pistol&rocket&skate board&table
    \tabularnewline
    \hline
Yi \cite{yi2016scalable}&81.4&81.0&78.4&77.7&75.7&87.6&61.9&\textbf{92.0}&85.4&82.5&\textbf{95.7}&\textbf{70.6}&91.9&\textbf{85.9}&53.1&69.8&75.3\\
3DCNN \cite{Qi_2017_CVPR}&79.4&75.1&72.8&73.3&70.0&87.2&63.5&88.4&79.6&74.4&93.9&58.7&91.8&76.4&51.2&65.3&77.1\\
PointNet \cite{Qi_2017_CVPR}&83.7&\textbf{83.4}&78.7&82.5&74.9&89.6&73.0&91.5&85.9&80.8&95.3&65.2&93&81.2&\textbf{57.9}&72.8&80.6\\
Kd-Net \cite{klokov2017escape}&77.2&79.9&71.2&80.9&68.8&88.0&72.4&88.9&\textbf{86.4}&79.8&94.9&55.8&86.5&79.3&50.4&71.1&80.2\\
\hline
Ours&
\textbf{84.3}&82.0&\textbf{84.3}&\textbf{86.0}&\textbf{76.9}&\textbf{89.9}&\textbf{73.9}&90.8&85.7&\textbf{82.6}&95.2&66.0&\textbf{94.0}&82.6&51.5&\textbf{73.5}&\textbf{81.8}\\
\hline
  
    \end{tabular}
    \caption{3DmFV-Net part segmentation performance compared to state of the art. The metric is mean IoU.}
    \label{table:SegAcc}
\end{table*}
    \begin{figure}
		\centering
		\includegraphics[width=0.3\textwidth]{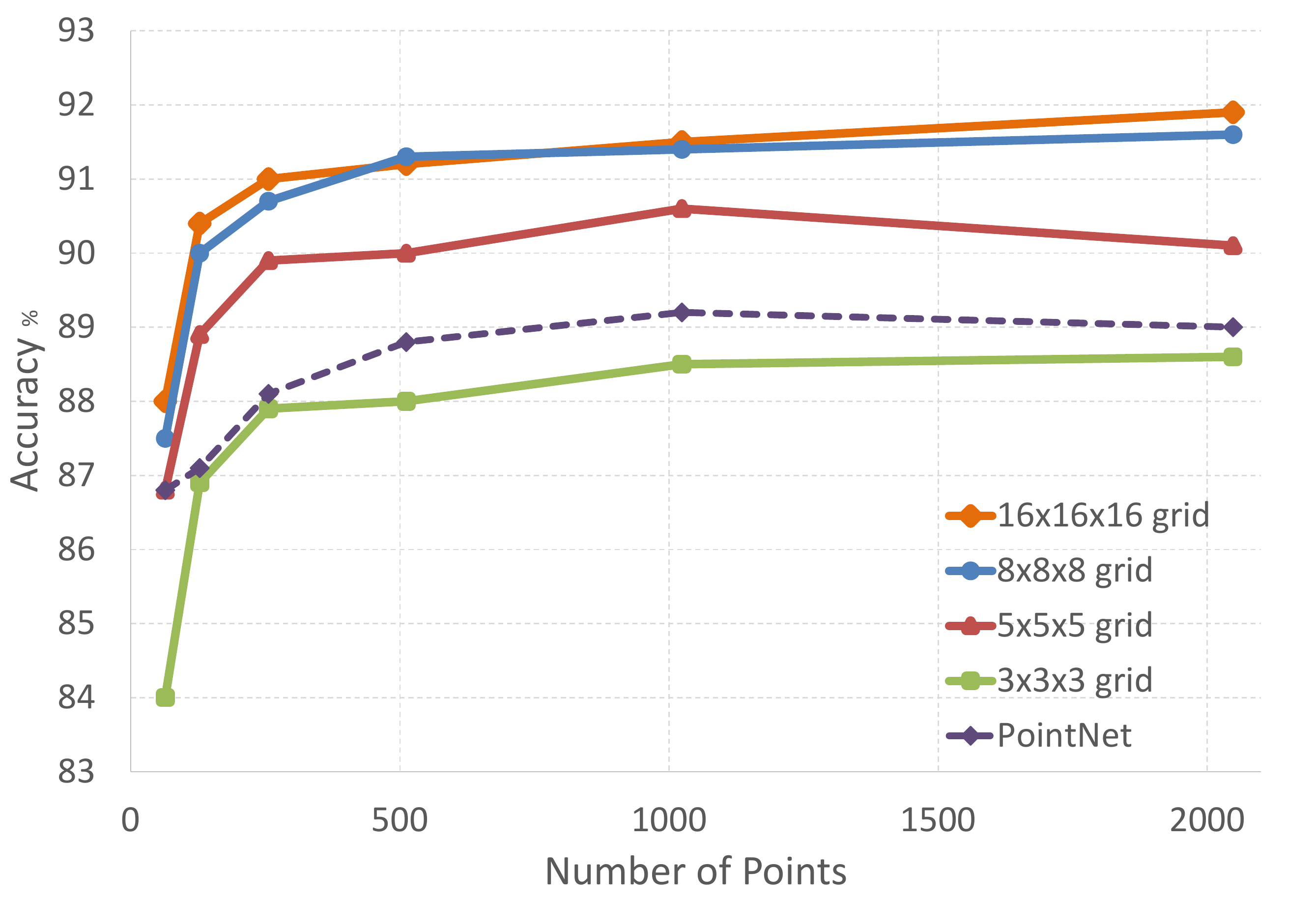}
		\caption{Effects of grid resolution and number of input points on the evaluation accuracy.}
		\label{fig:acc_vs_npoints}	
	\end{figure}
\textbf{Resolution and standard deviation}
We found that accuracy increases with both grid size ($m$) and the number of points representing the model; see Figure \ref{fig:acc_vs_npoints}. Note that performance saturates in both parameters. PointNet \cite{Qi_2017_CVPR} seems to be more sensitive to the number of points. This is probably because their descriptors are learned as well.
The architecture is slightly different for each grid size; see appendix \ref{appx:3DmFV_grid_arch}.

We also found that the model is insensitive to the selection of standard deviation as long as it is not too small. In the case of very small $\sigma$, most points do not contribute to any Gaussian, rendering the FV representation empty; See appendix \ref{appx:robustness}.
 \begin{figure}
\centering
\begin{tabular}{cc}	\includegraphics[width=0.22\textwidth]{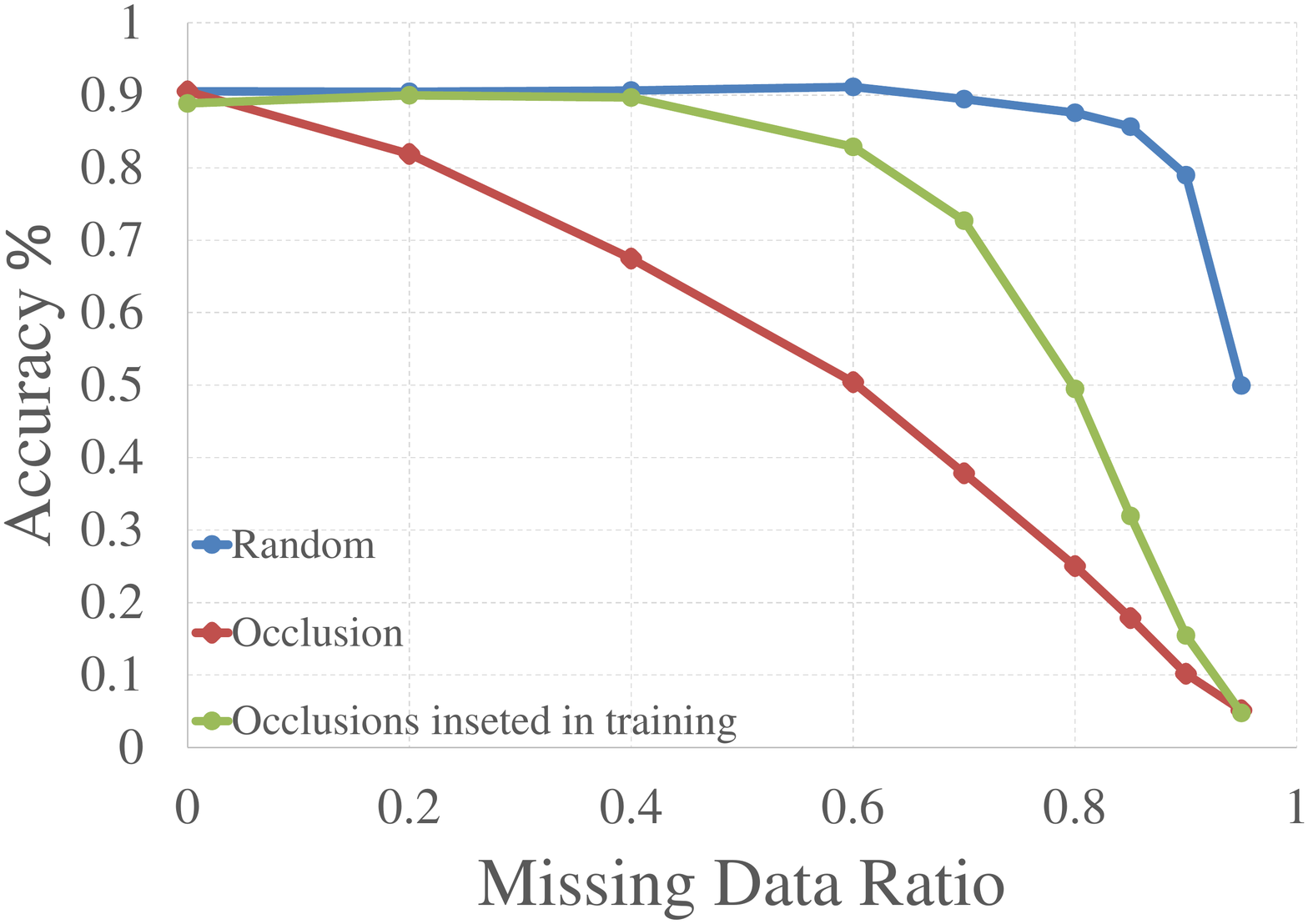}
&			
\includegraphics[width=0.22\textwidth]{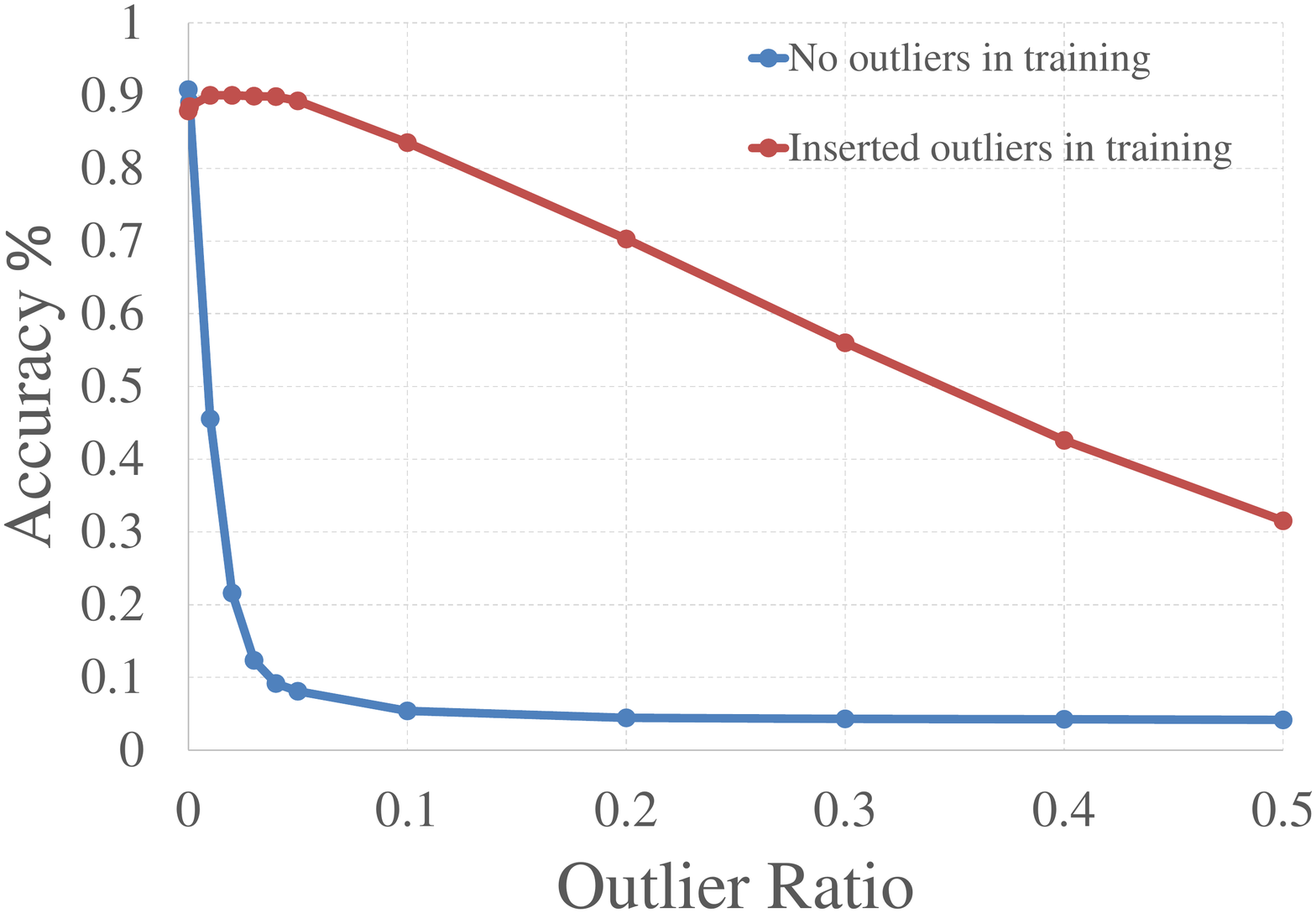}						
\\
\includegraphics[width=0.22\textwidth]{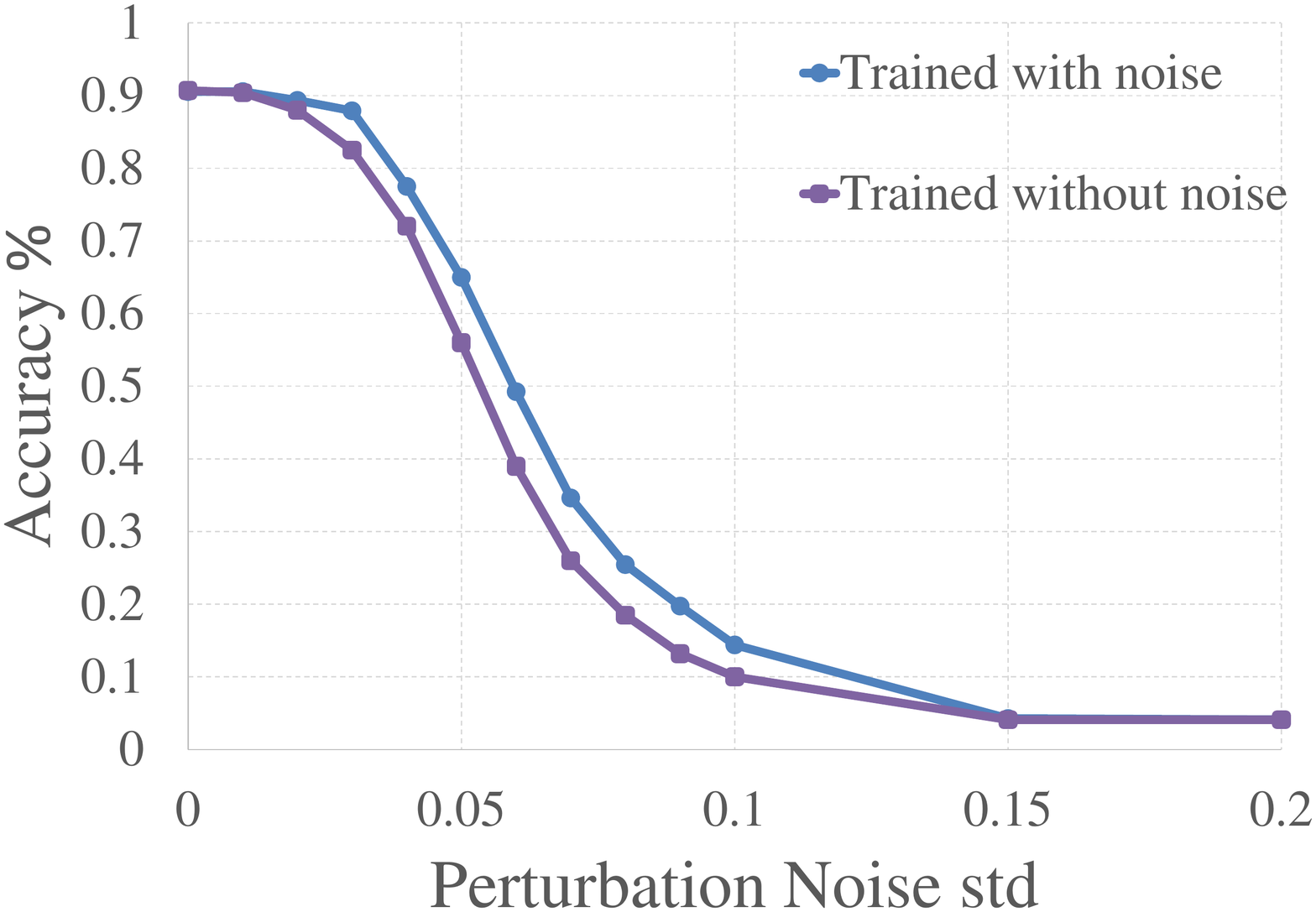}
&				
\includegraphics[width=0.22\textwidth]{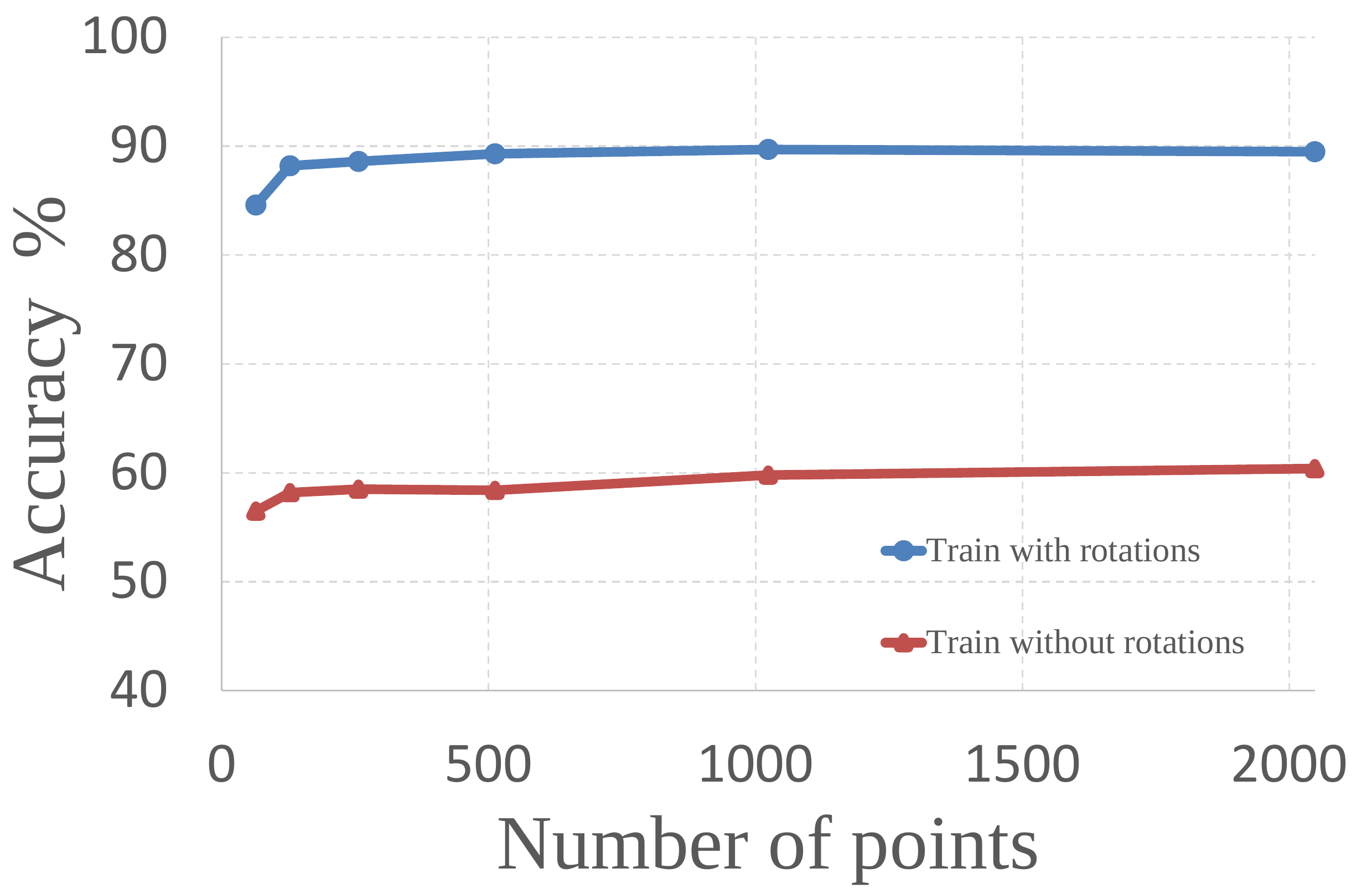}\\
	\end{tabular}			
\caption{3DmFV-Net robustness to data corruptions. Classification accuracy results for missing data (top-left), outlier insertion (top-right), perturbation noise (botom-left), and rotations (bottom-right).}
		\label{fig:robustness}	
	\end{figure} 
\subsection{Robustness evaluation}
 To simulate real-world point cloud classification challenges, we tested 3DmFV-Net's robustness under several types of noise:
%  \begin{enumerate}[label={(\alph*)}]
%	\item Uniform points deletion - randomly deleting points is equivalent to classifying clouds that are smaller than those used for training. .
%	\item Focused region points deletion - selecting a random point and deleting its closest points (the number of points is defined by a given ratio) mimicking occlusions.
%	\item Outlier points - adding uniformly distributed points. 
%	\item Perturbation noise  - Adding bounded Gaussian noise\MIC{What is this noise?} independetly to all points,  expressing measurement inaccuracy.\MIC{maybe we should use one verb out of expressingm mimicking,. perhaps simulating ?}
%	\item Random rotation -  randomly rotating the point cloud w.r.t. the global reference frame expressing the unknown scanned object orientation.
%\end{enumerate}  
 \\
	 \textbf{Uniform point deletion -} randomly deleting points is equivalent to classifying clouds that are smaller than those used for training.\\
	 \textbf{Focused region point deletion -} selecting a random point and deleting its closest points (the number of points is defined by a given ratio), simulating occlusions.\\
	 \textbf{Outlier points -} adding uniformly distributed points.\\ 
	 \textbf{Perturbation noise -} adding small translations, with a bounded Gaussian magnitude, independently to all points,  simulating measurement inaccuracy.\\
	\textbf{Random rotation -} randomly rotating the point cloud w.r.t. the global reference frame, simulating the unknown orientation of a scanned object.
  
The results (Figure  \ref{fig:robustness}) demonstrate that the proposed approach is inherently robust to perturbation noise and uniform point deletions. For the other types of data corruptions, training the classifiers with any of these types of noise made it robust to them.
\begin{figure}
\centering
		\includegraphics[width=0.48\textwidth]{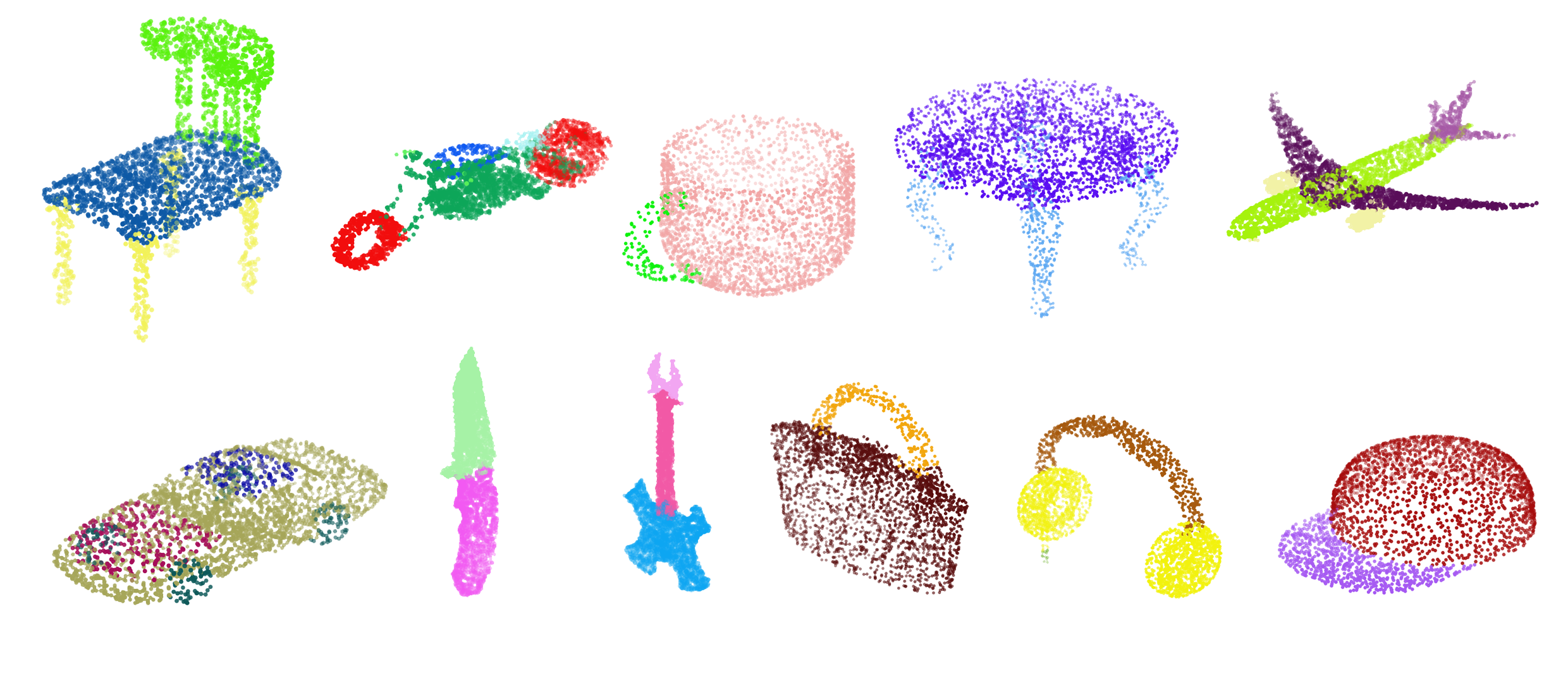}
		\caption{3DmFV-Net part segmentation qualitative results}
		\label{fig:seg_qualitative}	
\end{figure}  
\subsection{Failure cases}
Analyzing misclassifications, we found that most of them occur for similar class pairs: table-desk, dresser-night stand, and plant-flower pot, containing objects which are difficult to discriminate even for humans. See  appendix \ref{appx:fails}  for some typical failures and for the confusion matrix (computed on the test set). 	

\subsection{Part segmentation}
We evaluate the performance of our part segmentation architecture on the ShapeNet part dataset from \cite{yi2016scalable}. It contains 16881 point clouds with 50 annotated parts from 16 categories. 
%  (per point ground truth annotation). 
Learning this dataset is challenging because it is highly imbalanced. 
%  with respect to the number of point clouds for each category and for parts within categories.
The evaluation metric is mean intersection over union (IoU). It is calculated first for each part of each point cloud and is then averaged over 
%  part-IoU is computed (for parts available in that category). Next, these IoUs are averaged over 
each point cloud, and over all point clouds 
% resulting in a point-cloud IoU. The resulting IoUs are averaged over all point clouds 
in that category, yielding the category IoU. The total mean IoU is computed as a weighted average of all category IoUs, where the weights are the respective number of point clouds in that category. We report in Table \ref{table:SegAcc} the total mean IoU and the category IoUs and compare to other state of the art methods. We achieve best performance in 9/16 categories while no other method wins in more than 4/16. Additionally, we achieve best mean IoU. Qualitative part-segmentation results are presented in Figure \ref{fig:seg_qualitative}, and additional qualitative results are presented in the appendix \ref{appx:part_seg}.
  
\section{Conclusion}
\label{Sec:summary}

In this work, we propose a new unsupervised 3D point cloud representation, the 3D modified Fisher vector. It preserves the raw point cloud data while using a grid for structure. This allows the use of the proposed CNN architecture (3DmFV-Net).
%In this work, we propose a new deep neural network architecture for point cloud data using the proposed unsupervised 3D modified Fisher vector representation. Some representation of the raw point cloud data seems essential to overcome its inherent lack of structure. 

Representing data by non-learned features goes against the deep network principle that the best performance is obtained only by learning all components of the classifier using an end-to-end optimization. The proposed representation achieves state of the art results relative to all methods that use point cloud input, and therefore provides evidence that end-to-end learning is not always essential.  

%\begin{acknowledgements}
%This work was supported by Magnet Omek Consortium, Ministry of Industry and Trade, Israel.
%\end{acknowledgements}
\bibliographystyle{plain}  
\bibliography{references}

\clearpage
\newpage

\section{Appendix}
\label{Sec:appendix}
\subsection{Point reconstruction from 3DmFV representation}
\label{appx:reconstruction_dev}
In Section \ref{SubSec:FV-DNN} we provide an expression for a plane's parameters, given the FV representation of points sampled on it. This result holds in the asymptotic sense and under the assumption of uniform sampling. It holds for a single Gaussian. Here we prove this result, generalize it to grid GMM and verify it experimentally.  
%First we show the reconstruction of a single point, next we show the reconstruction of a line and finally a plane. 
%\textbf{A single Gaussian representing a single point} -
%In this case $T=1$,$K=1$. Here we  will use $w_k = w$ $\mu_k=\mu$, and $\sigma_k=\sigma$ for neatness. In addition, the Gaussian weight is  $w = 1$ and the soft assignment term becomes $\gamma_t(k) = \gamma_t(1) = 1$. this simplifies many expressions significantly since there is no summation. Therefore \cref{eq:dev_w,eq:dev_mu,eq:dev_sig} becomes:  
%
%	 \begin{equation} \label{eq:dev_mu_single_point_1_full}
%	 \begin{array}{l}
%	 \mathscr{G}_{w}^X = 0, \\
%	 	\mathscr{G}_{\mu}^X =  \frac{x_1-\mu}{\sigma}, \\
%	 	\mathscr{G}_{\sigma}^X = \frac{1}{\sqrt{2}} \left(\frac{ \left(x_1-\mu\right)^2}{\sigma^2}-1\right). 
%	 	\end{array}
%	 \end{equation}
%This is a system of six equations with three variables which yields a unique analytical expression to reconstruct the point from the 3DmFV representation with constraints on the representation which are satisfied inherently
% Therefore the solution is given by: 
%\begin{equation} \label{eq:x_single_point_full}
%\begin{array}{l}
%	 	x_1 = \sigma_k\mathscr{G}_{\mu_k}^X +\mu_k,\\
%	 	\mathscr{G}_{w}^X = 0, \\
%	 	\mathscr{G}_{\sigma}^X = \pm\sqrt{1+\sqrt{2}\mathscr{G}_{\mu}^X}.
%\end{array}
%\end{equation}
\\

\begin{figure}
		\centering
\centering
\begin{tabular}{cc}
\includegraphics[width=0.2\textwidth]{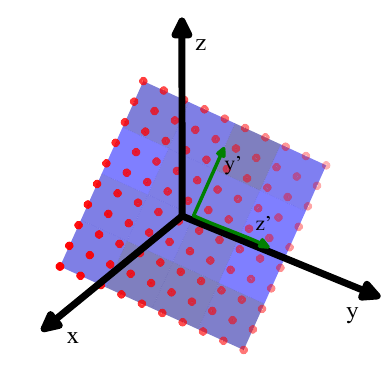}	&	\includegraphics[width=0.2\textwidth]{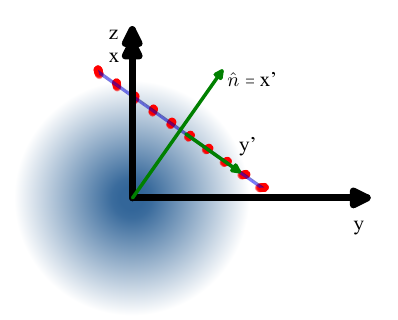}\\
\end{tabular}
\includegraphics[width=0.2\textwidth]{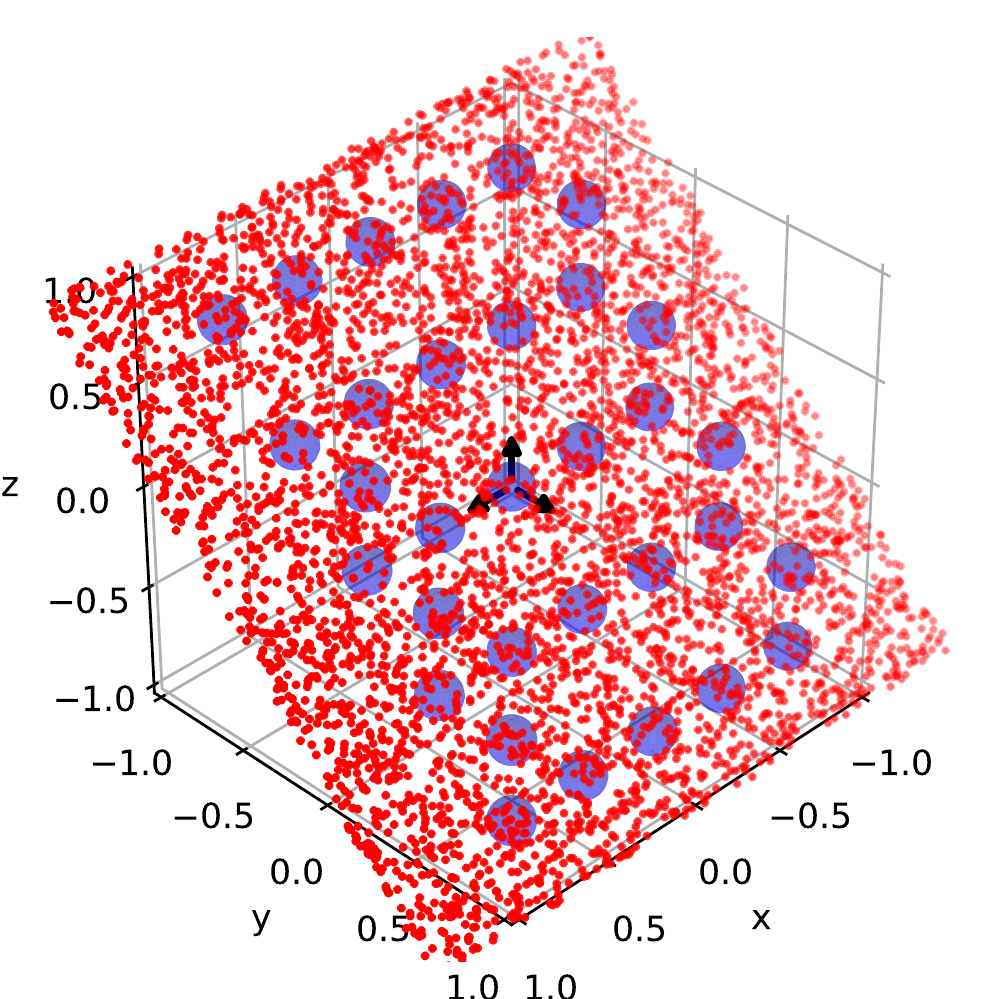}						
		\caption{Points on a plane with their reconstructed plane. A view from the $\hat{n}$ direction (top-left), a view from the $y'$ direction, centered in one of the Gaussians (top-right), and an isometric view Gaussians that have a non-negligible $\sum_{t=1}^{T}\gamma_t(k)$ value with points on the plane (bottom).}
		\label{fig:plane_reconstruction}	
	\end{figure}

The plane equation is given by $ \hat{n}^T\bm{p} = \rho$, where $\hat{n}= (a, b, c)^T$ is a unit  normal to the plane and $\rho$ is its distance from the origin. 
Using the assumption that $\gamma_t(k)$ is approximately sharply peaked \cite{sanchez2013image}, we consider the $k$-th Gaussian and the $T_k$ points for which  $\gamma_t(k) \approx 1$. From eq. \ref{eq:dev_w} we get an expression for $T_k$:
\begin{equation}
	T_k \approx \sum_{t=1}^{T}\gamma_t(k) = T\sqrt{w}(\mathscr{G}_{\alpha_k}^X + \sqrt{w})
\end{equation}
\\
For this Gaussian, eq. \ref{eq:dev_mu} is  simplified to: 
\begin{equation} \label{eq:g_mu_peaked}
\mathscr{G}_{\mu}^X = \frac{T_k}{\sigma\sqrt{w}} \sum_{t=1}^T (\bm{p}_t - \mu).
\end{equation}
The $\mathscr{G}_{\mu}^X $ expression becomes clearer when we change the coordinate system to $x'y'z'$ (see Figure  \ref{fig:plane_reconstruction} top-right), for which the  origin is at the Gaussian center, the axis  $x'$ coincides with $\hat{n}$ and the other axes are chosen to constitute an orthogonal system.  We select: 
\begin{equation}
\hat{x}'=\hat{n}, 
\hat{y}' = \frac{\hat{n} \times \hat{y}}{\left\lvert \hat{n}\times \hat{y} \right\lvert},
 \hat{z}' = \frac{\hat{y}'\times \hat{n}}{\left\lvert \hat{y}' \times \hat{n} \right\lvert} .
\end{equation}
This yields the following transformation:
\begin{equation} \label{eq:transform_p}
\bm{p} = 
\left( \begin{matrix}
   x  \\
   y  \\
   z  \\
\end{matrix}\right) 
= \mu +
 \left( \begin{matrix}
   a & -\frac{c}{\sqrt{{{a}^{2}}+{{c}^{2}}}} & -\frac{ab}{\sqrt{{{a}^{2}}+{{c}^{2}}}}  \\
   b & 0 & \sqrt{{{a}^{2}}+{{c}^{2}}}  \\
   c & \frac{a}{\sqrt{{{a}^{2}}+{{c}^{2}}}} & -\frac{bc}{\sqrt{{{a}^{2}}+{{c}^{2}}}}  \\
\end{matrix} \right)\left( \begin{matrix}
   x'  \\
   y'  \\
   z'  \\
\end{matrix} \right).
\end{equation}
For all points on the plane, $x' = \rho_0$. 
By symmetry, for random uniform sampling on the plane, the expected value of both $y'$ and $z'$ is zero. Therefore, for a large number of samples, inserting  eq. \ref{eq:transform_p} in eq. \ref{eq:g_mu_peaked} yields 
\begin{equation} \label{eq:g_mu_approx}
\mathscr{G}_{\mu}^X \approx  \frac{T_k}{\sigma\sqrt{w}} 
\left( \begin{matrix}
   a  \\
   b  \\
   c  \\
\end{matrix}\right)
x'.
\end{equation}
In addition, recall that to remove the dependency on the total number of points $T$, we divide by $T$ (eq. \ref{eq:FV_norm_T}). 
A simple inversion of eq. \ref{eq:g_mu_approx} yields: 
\begin{equation}
  a = \mathscr{G}_{\mu_x}^X\frac{\sigma\sqrt{w}T}{T_k\rho_0 } ,
  b = \mathscr{G}_{\mu_y}^X\frac{\sigma\sqrt{w}T}{T_k\rho_0},
  c = \mathscr{G}_{\mu_z}^X\frac{\sigma\sqrt{w}T}{T_k\rho_0}.
\end{equation}
Using the constraint that  $\hat{n}$ is a unit vector, 
\begin{equation}
a^2+b^2+c^2=1, 
\end{equation}
we derive the expressions for the plane parameters as a function of (some of) the FV components:
\begin{equation} \label{eq:plane_reconstruction}
	a = \frac{\mathscr{G}_{\mu_x}}{\left\lVert \mathscr{G}_{\mu} \right\rVert},
	b = \frac{\mathscr{G}_{\mu_y}}{\left\lVert \mathscr{G}_{\mu} \right\rVert},
	c = \frac{\mathscr{G}_{\mu_z}}{\left\lVert \mathscr{G}_{\mu} \right\rVert},
	\rho_0 =  \frac{\sigma \left\lVert \mathscr{G}_{\mu} \right\rVert}{\mathscr{G}_{\alpha_k}^X+\sqrt{w}}. 
\end{equation}
Note that $\rho_0$ is the distance of the plane along $\hat{n}$ in the local coordinate system. Therefore, to compute $\rho$ in the global coordinate system we use:
\begin{equation} \label{eq:plane_reconstruction_rho}
\rho = \rho_0 + \mu^T \hat{n}
\end{equation}% Here , $	\left\lVert \mathscr{G}_{\mu} \right\rVert = \sqrt{\mathscr{G}_{\mu_x}^2 + \mathscr{G}_{\mu_y}^2 + \mathscr{G}_{\mu_z}^2}$.

To validate these expressions in a more realistic case, where $\gamma_t(k)$ is not binary, we sampled points uniformly on a plane specified by $\frac{1}{\sqrt{2}}(0, 1, 1)\bm{p} = 0.05$. We then computed the point cloud's FV representation using a 5$\times$5$\times$5 Gaussian grid ($\sigma=\frac{1}{10}, w=\frac{1}{125}$) using eq. \ref{eq:dev_w} to \ref{eq:dev_sig}. Next, using eq. \ref{eq:plane_reconstruction} and \ref{eq:plane_reconstruction_rho} we estimated the plane parameters from the FV representation of each Gaussian that has a non-negligible $\sum_{t=1}^{T}\gamma_t(k)$ value, see \ref{fig:plane_reconstruction} (bottom). Figure \ref{fig:plane_reconstruction} (top-left) shows a part of the original sampled points in red and the reconstructed surface in purple. We found that all the reconstructed planes are similar to the original one,  which demonstrate that the FV components hold the information about the plane parameters. 
%As expected, the accuracy improves with increasing number of points and we observed that reducing the $\sigm$ value makes the sharply peaked assumption...   
\subsection{Failure cases}
\label{appx:fails}
\begin{figure}
		\centering
		\includegraphics[width=0.48\textwidth]{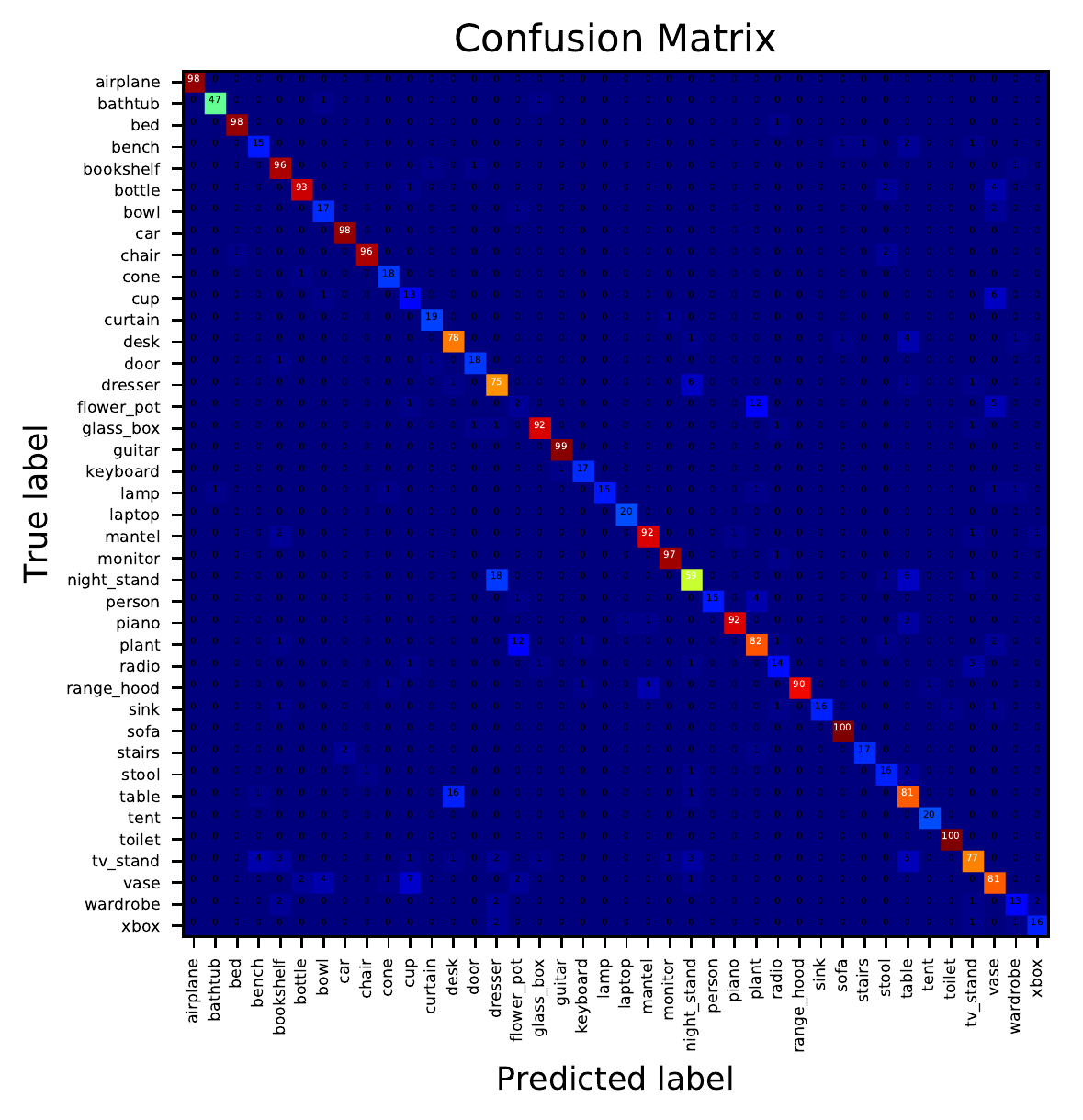}
		\caption{3DmFV-Net classification confusion matrix}
		\label{fig:confusion_mat}	
	\end{figure}

\begin{figure}[t]
		\centering
		\includegraphics[width=0.48\textwidth]{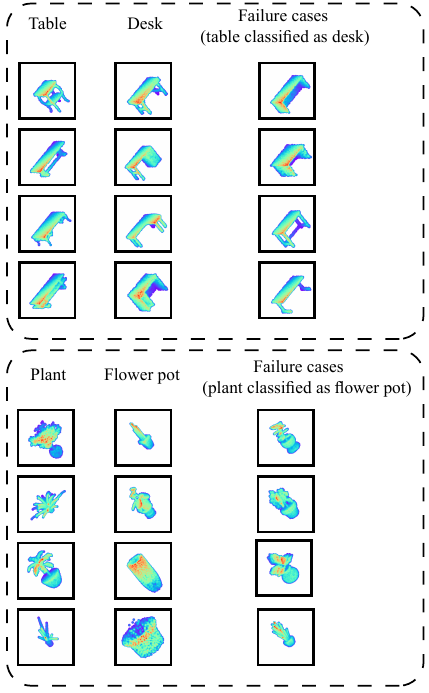}
		\caption{3DmFV-Net classification failure cases. Table point clouds classified as desks (top), and plants classified as flower pots (bottom).}
		\label{fig:fail_cases}	
	\end{figure}

In order to better understand the 3DmFV representation and the 3DmFV-Net, we explored failure cases. The confusion matrix in Figure \ref{fig:confusion_mat} shows that the majority of misclassified point clouds belong to a few class pairs, specifically table-desk, plant-flowerpot and dresser-night stand. Further inspection using visualization of the failure cases presented in Figure \ref{fig:fail_cases} provides some insight. First, some classes are inherently hard to distinguish even for humans (table-desk) and second, the training data imposes a challenge since there is some overlap between categories (plant-flower pot). 
	
\subsection{3DmFV architecture for different grid sizes}
\label{appx:3DmFV_grid_arch}
We tested several grid resolutions for the 3DmFV representation (see Section \ref{Sec:results}, Figure \ref{fig:acc_vs_npoints}) . Each grid size imposes a slightly different network architecture. The architectures are detailed below. See Figure \ref{fig:inception_module} for a description of the parametric inception module. 
% Note that the notation  grid $m$ refers to an $m\times m\times m$ grid
\begin{enumerate}

\item Grid 3$\times$3$\times$3: 3DmFV-inception(2,3,64) - inception(2,3,128) - inception(2,3,256) - inception(2,3,256) - inception(2,3,512) - FC(1024) - FC(256) - FC(128) - FC(\#classes)

\item Grid 5$\times$5$\times$5: 3DmFV - inception(3,5,64) - inception(3,5,128) - inception(3,5,256) - maxpool([3,3,3],2) - inception(2,3,256) - inception(2,3,512) - FC(1024) - FC(256) - FC(128) - FC(\#classes)

\item  Grid 8$\times$8$\times$8 (3DmFV-Net): 3DmFV - inception(3,5,64) - inception(3,5,128) - inception(3,5,256) - maxpool([2,2,2],2) - inception(3,5,256) - inception(3,5,512) - maxpool([2,2,2],2) - FC(1024) - FC(256) - FC(128) - FC(\#classes)

\item Grid 16$\times$16$\times$16 : 3DmFV - inception(4,8,64) - inception(4,8,128) - inception(4,8,256) - maxpool([2,2,2],2) - inception(3,5,256) - inception(3,5,512) - maxpool([2,2,2],2) - inception(2,3,512) - inception(2,3,512) - maxpool([2,2,2],2) - FC(1024) - FC(256) - FC(128) - FC(\#classes)
\end{enumerate}
\subsection{Additional robustness tests}
\label{appx:robustness}
\begin{figure}
		\centering
		\includegraphics[width=0.35\textwidth]{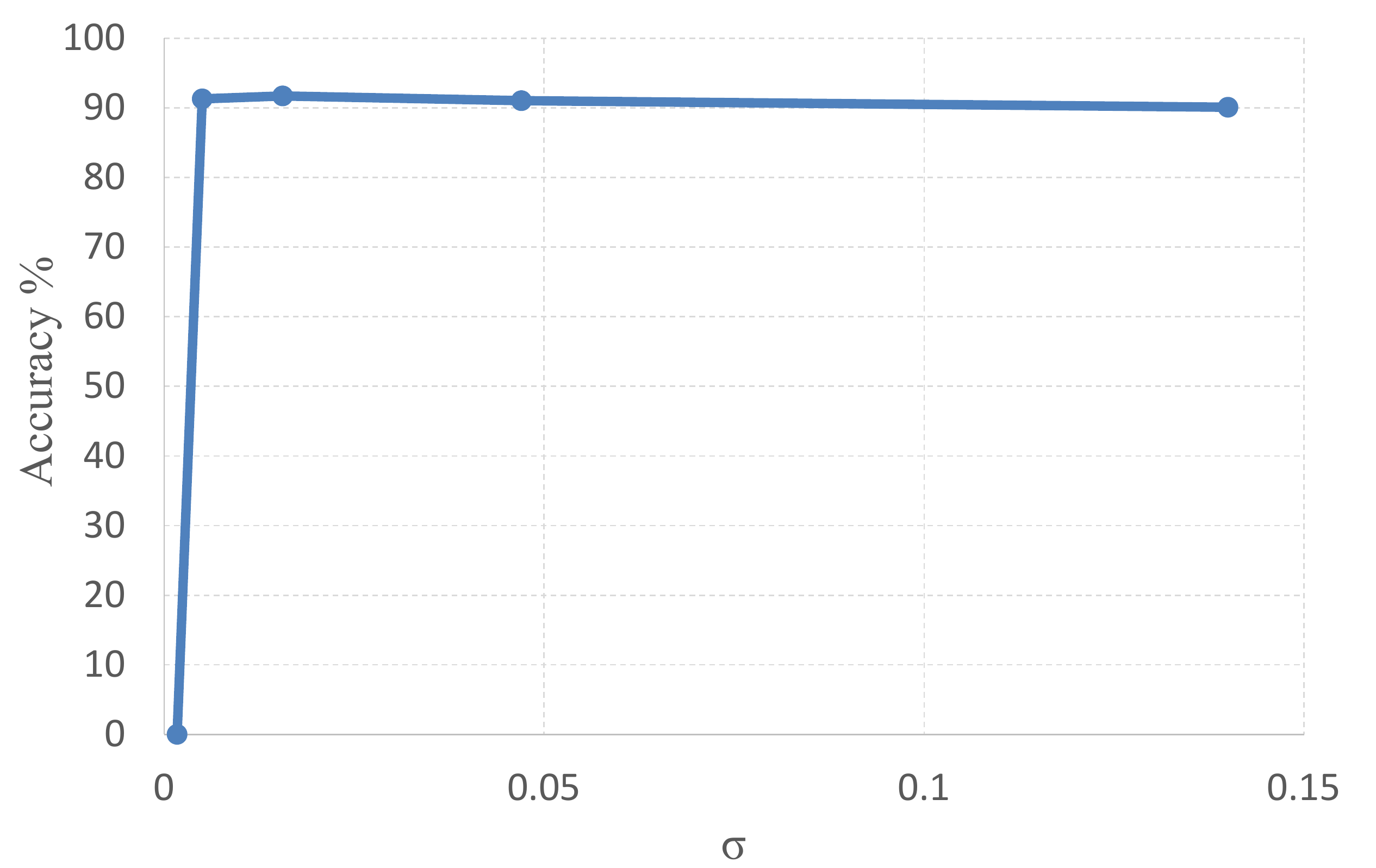}
		\caption{Effects of Gaussian standard deviation ($\sigma$) selection on the evaluation accuracy}
		\label{fig:acc_vs_sigma}	
	\end{figure}

 One of the parameters of the 3DmFV representation is the Gaussian standard deviation $\sigma$. The value of $\sigma$ qualitatively specifies spherical sub-volume whose points contribute to the 3DmFV component associated with a specific Gaussian. Very small $\sigma$ values create Gaussians with very few or no contributing points and very large $\sigma$s create Gaussians that may be affected by many or all points in the cloud. We tested the robustness of 3DmFV-Net to the $\sigma$ parameter selection. Figure \ref{fig:acc_vs_sigma} shows that the network is robust to the $\sigma$ selection except for very small $\sigma$s, for which the network performs poorly (since the representation essentially fails to capture the point cloud).   
  
\subsection{Part segmentation}
\label{appx:part_seg}

\begin{figure}
		\centering		\includegraphics[width=0.43\textwidth]{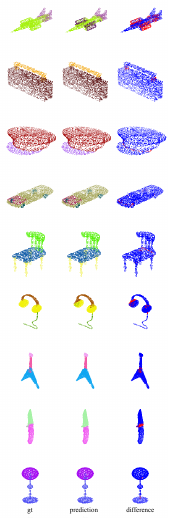}
		\caption{Part segmentation qualitative comparative results. Ground truth labels (left), predicted labels (middle), and color coded difference (right).}
		\label{fig:segmentation_diff}	
	\end{figure}

In Section \ref{SubSec:3DmFVNet_seg} we presented qualitative results for part segmentation. Additional part segmentation results are presented here with a comparative visualization. In Figure \ref{fig:segmentation_diff}, the left column shows the ground truth point labels, the middle column shows the  labels predicted by the 3DmFV-Net segmentation network, and the right column shows a color coded comparison between the two, where correctly labeled points are shown in blue and mislabeled in red. It can be seen that mislabeled points sometimes appear in transitional locations between labels (e.g., red points are visible where the chair back meets the chair seat).    

\end{document}